\PassOptionsToPackage{table}{xcolor}
\documentclass[10pt,twocolumn,letterpaper]{article}

\usepackage[pagenumbers]{cvpr} % To force page numbers, e.g. for an arXiv version

\definecolor{cvprblue}{rgb}{0.21,0.49,0.74}
\usepackage[pagebackref,breaklinks,colorlinks,allcolors=cvprblue]{hyperref}

\usepackage{hyperref} 
\hypersetup{
	colorlinks=true,
	linkcolor=red,
	filecolor=red,      
	urlcolor=magenta,
	citecolor=blue
}
\usepackage{amsmath}
\usepackage{amsfonts}
\usepackage{booktabs} 
\usepackage{multicol}
\usepackage{multirow}
\usepackage{placeins}
\usepackage{marvosym}
\usepackage{wrapfig}
\usepackage{algorithm}
\usepackage{courier}
\usepackage[table]{xcolor}
\def\paperID{2969} % *** Enter the Paper ID here
\def\confName{CVPR}
\def\confYear{2025}

\title{\textcolor{blue}{OverLoCK}: An \textcolor{blue}{Over}view-first-\textcolor{blue}{Lo}ok-\textcolor{blue}{C}losely-next ConvNet\\ with Context-Mixing Dynamic \textcolor{blue}{K}ernels}

\author{Meng Lou 
\and
Yizhou Yu
\and
School of Computing and Data Science, The University of Hong Kong \\
{\tt\small loumeng@connect.hku.hk}, {\tt\small yizhouy@acm.org}
}

\begin{document}
\maketitle

\begin{abstract}
%The human vision system leverages top-down attention to achieve accurate perception, characterized by an ``overview first, look closely next" mechanism where coarse scene analysis guides finer-grained examination. 
Top-down attention plays a crucial role in the human vision system, wherein the brain initially obtains a rough overview of a scene to discover salient cues (i.e., overview first), followed by a more careful finer-grained examination (i.e., look closely next).
However, modern ConvNets remain confined to a pyramid structure that successively downsamples the feature map for receptive field expansion, neglecting this crucial biomimetic principle. We present OverLoCK, the first pure ConvNet backbone architecture that explicitly incorporates a top-down attention mechanism. Unlike pyramid backbone networks, our design features a branched architecture with three synergistic sub-networks: 1) a Base-Net that encodes low/mid-level features; 2) a lightweight Overview-Net that generates dynamic top-down attention through coarse global context modeling (i.e., overview first); and 3) a robust Focus-Net that performs finer-grained perception guided by top-down attention (i.e., look closely next). To fully unleash the power of top-down attention, we further propose a novel context-mixing dynamic convolution (ContMix) that effectively models long-range dependencies while preserving inherent local inductive biases even when the input resolution increases, addressing critical limitations in existing convolutions. Our OverLoCK exhibits a notable performance improvement over existing methods. For instance, OverLoCK-T achieves a Top-1 accuracy of 84.2\%, significantly surpassing ConvNeXt-B while using only around one-third of the FLOPs/parameters. On object detection, our OverLoCK-S clearly surpasses MogaNet-B by 1\% in AP$^b$. On semantic segmentation, our OverLoCK-T remarkably improves UniRepLKNet-T by 1.7\% in mIoU. 
Code is publicly available at \url{https://bit.ly/OverLoCK}.
\vspace{-10pt}
\end{abstract}

\section{Introduction}
\label{sec:intro}
% \vspace{-10pt}
Top-down neural attention~\cite{saalmann2007neural,gilbert2007brain,li2014understanding} is a crucial perception mechanism in the human vision system, which suggests that the brain initially processes a visual scene to quickly form an overall high-level perception, which goes back to fuse with the sensory input, enabling the brain to make more accurate judgments, such as object locations, shapes, and categories. Many previous works have incorporated such top-down attention into vision models, but some of them are unsuitable for building modern vision backbones due to incompatible model designs~\cite{hu2016bottom,xu2016ask,anderson2018bottom,mittal2020learning,chen2020blendmask} while the remaining methods primarily focus on recurrent architectures~\cite{cao2015look,zamir2017feedback,cao2018feedback,pang2021tdaf,shi2023top}, which introduce additional computational overhead due to recurrent operations, resulting in a suboptimal trade-off between performance and computational complexity.
\par
\begin{figure}[t]
    \centering
    \includegraphics[width=0.475\textwidth]{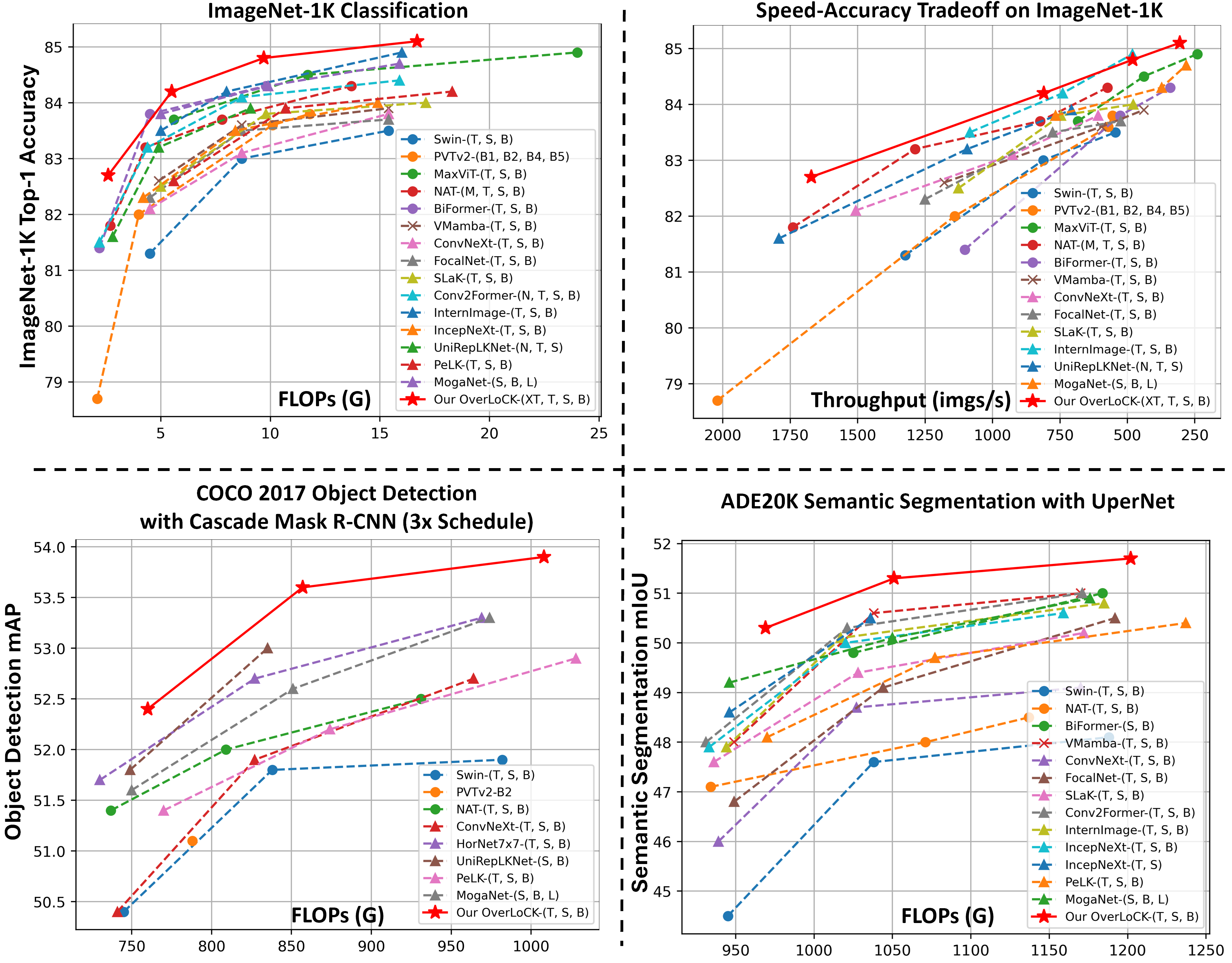}
    \caption{Performance comparisons between our OverLoCK and other representative backbone networks on vision tasks.}
    \label{fig:acc_plot}
    \vspace{-15pt}
\end{figure}
\begin{figure}[t]
\centering
\includegraphics[width=0.49\textwidth]{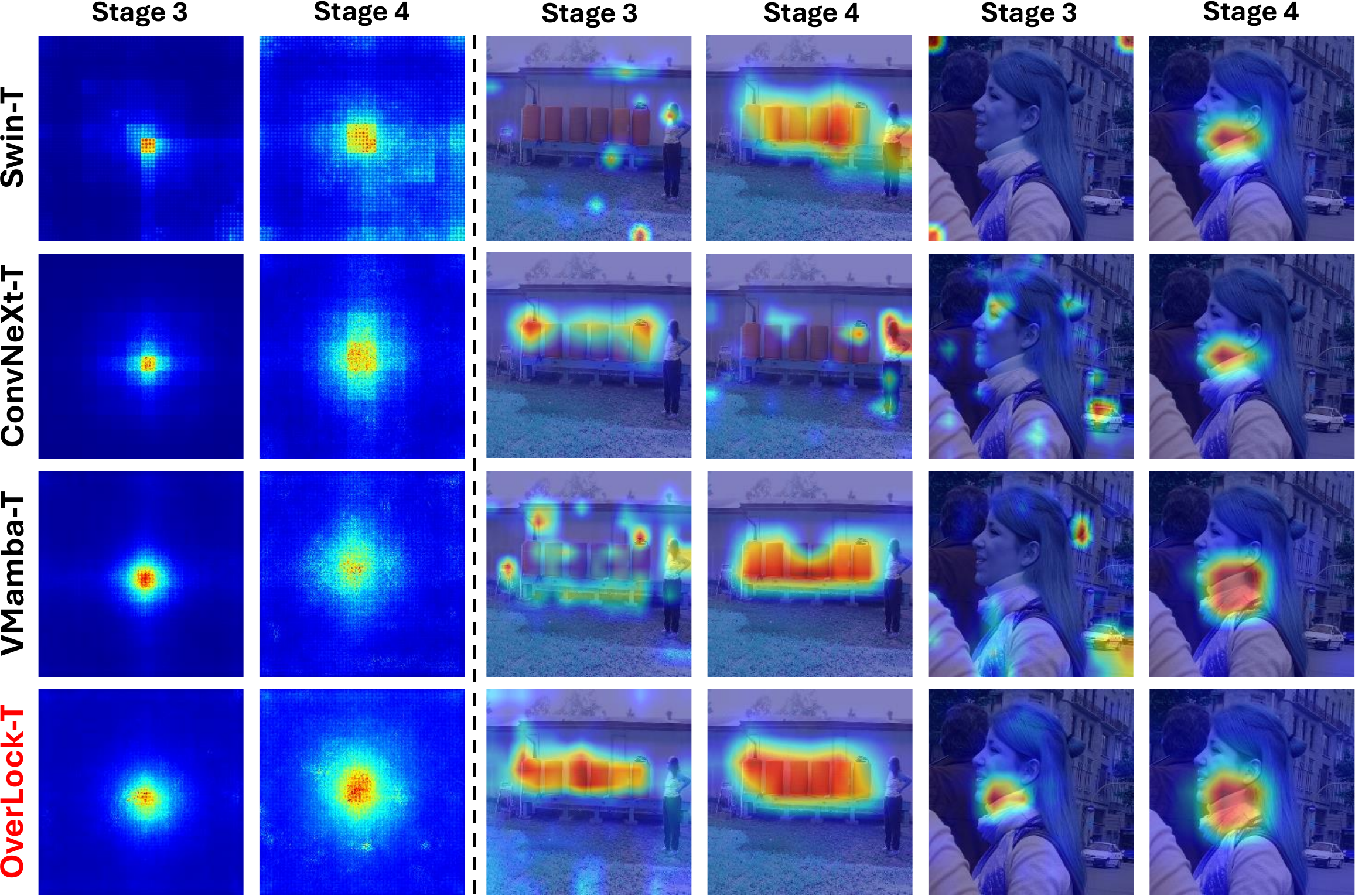}
\vspace{-17.5pt}
\caption{
\textbf{(a)} Comparison of Effective Receptive Fields (ERF)~\cite{luo2016understanding} at the last layer of deep stages (i.e., Stages 3 and 4) among backbone networks. The results are obtained by averaging over 300 images from ImageNet-1K validation set. As shown, despite being a pure ConvNet, OverLoCK-T has a larger ERF than VMamba-T that emphasizes global modeling, in both Stages 3 and 4. \textbf{(b)} Visualizations of class activation maps computed using Grad-CAM~\cite{selvaraju2020grad} for the output of deep stages (i.e., Stages 3 and 4). The category labels of these two images are ``Barrel" and ``Neck Brace". The results demonstrate that although classic hierarchical models can capture long-range dependencies to varying degrees, they struggle to localize objects with the correct category label, especially in Stage 3, which is farther from the classifier. In contrast, our proposed new network architecture can produce more accurate class activation maps in both Stages 3 and 4.
}
\label{fig:erf_intro}
\vspace{-19.5pt}
\end{figure}

A key property of the top-down attention mechanism is the use of feedback signals as explicit guidance to locate meaningful regions in a scene \cite{saalmann2007neural}. However, the classic hierarchical architecture employed in most existing vision backbones~\cite{he2016deep,liu2021swin,wang2021pyramid,wang2022pvt,liu2022convnet,liu2024vmamba} contrasts with this biological mechanism, as it progressively encodes features from lower to higher levels so that the input features of a layer rely solely on features from previous layers. Hence, there is a lack of explicit top-down semantic guidance in the operations at intermediate layers. To investigate this, we visualize the class activation maps~\cite{selvaraju2020grad} and the effective receptive fields (ERFs)~\cite{luo2016understanding} of three representative hierarchical vision models. Swin-T~\cite{liu2021swin}, ConvNeXt-T~\cite{liu2022convnet}, and VMamba-T~\cite{liu2024vmamba}. As shown in Figure~\ref{fig:erf_intro}, these image classification models struggle to accurately localize objects with the correct category label in the feature maps, especially in Stage 3, which is farther from the classifier layer, despite capturing long-range dependencies in varying degrees. Therefore, \textit{how to develop a modern ConvNet that leverages the top-down attention mechanism while achieving an excellent performance-complexity trade-off remains an open problem}.

On the basis of the above discussion, we propose a biomimetic Deep-stage Decomposition Strategy (DDS) inspired by the top-down attention mechanism in the human vision system. Unlike previous works, our goal is to enhance both feature maps and kernel weights in ConvNets with guidance from dynamic top-down semantic contexts. As illustrated in Figure~\ref{fig:net}, DDS decomposes a network into three sub-networks: Base-Net, Overview-Net, and Focus-Net. Specifically, a Base-Net encodes low-level and mid-level information, the output is fed into a lightweight Overview-Net to rapidly gather a semantically meaningful but low-quality context representation, analogous to the ``overview" process in visual perception. Subsequently, we designate the output of the Overview-Net as a context prior, which, along with the output of the Base-Net, is fed into a deeper and more powerful Focus-Net to obtain more accurate and informative high-level representations, akin to the ``look closely" process in visual perception.
\par
As a top-down context contains information across the entire input image, to fully unleash its power and absorb its information into convolution kernels, the Focus-Net should utilize a powerful dynamic convolution as the token mixer, capable of adaptively modeling long-range dependencies to produce large receptive fields while preserving local inductive biases to capture nuanced local details. Nonetheless, we find that existing convolutions cannot meet these requirements simultaneously. Unlike self-attention mechanisms~\cite{dosovitskiy2020image,wang2021pyramid,wang2022pvt,tu2022maxvit,zhu2023biformer} and State Space Models~\cite{gu2023mamba,zhu2024vision,liu2024vmamba,lou2024sparx,fu2025segman} that can adaptively model long-range dependencies at various input resolutions, large kernel convolutions \cite{liu2022convnet,ding2022scaling,liu2022more,yu2023inceptionnext,ding2023unireplknet} and dynamic convolutions \cite{chen2020dynamic,li2021involution,li2022omni} are still confined to finite regions due to fixed kernel sizes even when input images have an increasingly large resolution, suggesting a weak long-range modeling ability. Although deformable convolutions \cite{dai2017deformable,wang2022internimage} can alleviate these issues to a certain extent, a deformable kernel shape sacrifices the inherent inductive bias of convolutions, giving rise to a relatively weak local perception ability. Hence, \textit{enabling pure convolutions to possess dynamic global modeling capabilities comparable to those of Transformer- and Mamba-based models while preserving strong inductive biases remains a challenge.}
\par
To tackle this problem, we introduce a novel \textbf{Cont}ext-\textbf{Mix}ing Dynamic Convolution (ContMix) that dynamically models long-range dependencies while maintaining strong inductive biases. Specifically, for every token in the input feature map, we compute its affinity with respect to a set of region centers across the top-down context feature map, yielding an affinity map. Subsequently, we utilize a learnable linear layer to transform every row of the affinity map, generating spatially varying dynamic convolution kernels. In this regard, every kernel weight carries global information from the top-down semantic context. Consequently, during convolution operations using our dynamic convolution kernels, each token interacts with the global information encoded in the kernels, thereby capturing long-range dependencies despite the fixed size of convolution kernels.
\par
Equipped with the proposed DDS and ContMix, we propose a novel \textbf{Over}view-first-\textbf{Lo}ok-\textbf{C}losely-next ConvNet with context-mixing dynamic \textbf{K}ernels (OverLoCK). As shown in Figure \ref{fig:acc_plot}, our OverLoCK demonstrates superior performance in comparison to representative ConvNet-, Transformer-, and Mamba-based models while striking an excellent balance between speed and accuracy. For example, on the ImageNet-1K dataset, OverLoCK-T achieves a Top-1 accuracy of 84.2\%, outperforming UniRepLKNet-T \cite{ding2023unireplknet} by 1\% and surpassing VMamba-T \cite{liu2024vmamba} by 1.6\%. On downstream tasks, OverLoCK also demonstrates leading performance. For example, OverLoCK-S outperforms MogaNet-B~\cite{li2023moganet} by 1.2\% in mIoU on semantic segmentation and surpasses PeLK-S \cite{chen2024pelk} by 1.4\% in AP$^b$ on object detection. Additionally, our method is capable of generating a larger ERF with a strong local inductive bias and more reasonable feature responses in comparison to other competitors, as shown in Figure~\ref{fig:erf_intro}.

\section{Related Work}
\label{sec:related_work}
\textbf{Evolution of ConvNets}. Since the debut of AlexNet~\cite{krizhevsky2012imagenet}, ConvNets gradually became the dominant architecture in computer vision. VGGNet~\cite{simonyan2014VGG} introduced the concept of stacking small kernels to build deep networks. ResNet~\cite{he2016deep} and DenseNet~\cite{huang2017densely} further proposed skip-connections to address the gradient vanishing/exploding issues in deep networks. However, with the rise of Vision Transformers~\cite{dosovitskiy2020image,touvron2021training,wang2021pyramid,liu2021swin,zhu2023biformer,lou2023transxnet}, ConvNets' dominance in vision tasks has been challenged. Hence, recent methods have proposed increasingly larger kernel sizes to mimic the self-attention mechanism~\cite{dosovitskiy2020image} and establish long-range dependencies~\cite{liu2022convnet,ding2022scaling,liu2022more,chen2024pelk,ding2023unireplknet,yu2023inceptionnext,woo2023convnext}. ConvNeXt~\cite{liu2022convnet} pioneered the use of 7$\times$7 kernels to build vision backbones, surpassing the performance of Swin Transformer~\cite{liu2021swin}. RepLKNet~\cite{ding2022scaling} further explored the promising performance of very large kernels by using 31$\times$31 kernels. On the other hand, gated mechanisms have been extensively explored in ConvNets~\cite{yang2022focalnet, rao2022hornet, li2023moganet, ma2024starnet, ma2024efficient}. For instance, MogaNet~\cite{li2023moganet} introduced a multi-order gated aggregation module to enhance the capacity for refining multi-scale feature representations. StarNet~\cite{ma2024starnet} unveiled the underlying reason for the superior performance of element-wise multiplications in the gated mechanism. More recently, RDNet~\cite{kim2024densenets} rethought the design of DenseNet and proposed an efficient densely-connected ConvNet. Unlike previous work, this paper focuses on improving the performance of ConvNets from both the architectural and mixer perspectives.
\par
\textbf{Dynamic Convolutions}. Dynamic convolution has been demonstrated to be effective in improving the performance of ConvNets~\cite{yang2019condconv,he2019dynamic,chen2020dynamic,li2022omni} by enhancing feature representation with input-dependent filters. Beyond regular channel-varying modeling, some methods~\cite{ramachandran2019stand,li2021involution,yuan2022volo,hassani2023neighborhood} have also proposed spatially varying modeling, which can generate distinct convolution weights for individual pixels in a feature map. Moreover, to enable both the weights and shape of convolution kernels to change dynamically, InternImage~\cite{wang2022internimage} re-designed deformable convolutions~\cite{dai2017deformable}, achieving notable performance gains. However, previous works have failed to simultaneously model long-range dependencies while preserving a strong local inductive bias, a limitation that our new dynamic convolution effectively addresses.
\par
% \textbf{Top-down Networks}. 
\textbf{Biomimetic Vision Models}. The human vision system has inspired the design of many excellent vision backbone networks. For instance, several advanced vision backbones~\cite{yang2022focalnet,min2022peripheral,chen2024pelk} have been inspired by the peripheral perception mechanism~\cite{lettvin1976seeing}, achieving notable performance. Likewise, the top-down attention mechanism~\cite{saalmann2007neural,gilbert2007brain,li2014understanding} has promoted developments in computer vision and machine learning, such as enhancing performance on specific tasks~\cite{hu2016bottom,xu2016ask,anderson2018bottom,mittal2020learning,chen2020blendmask}, exploring new learning algorithms~\cite{zhang2018top}, and designing generic architectures with a recurrent style~\cite{cao2015look,zamir2017feedback,cao2018feedback,pang2021tdaf}. Recently, AbsViT~\cite{shi2023top} introduced a feedback-based Vision Transformer backbone that reuses network outputs to recalibrate early features. In contrast to the above works, we propose a novel modern ConvNet-based vision backbone network that can efficiently generate and utilize top-down guidance, achieving significant performance gains across diverse vision tasks.

\section{Methodology}
\label{sec:method}
\subsection{Deep-stage Decomposition}
\label{sec:dds}
\begin{figure}[!t]
\centering
\includegraphics[width=0.49\textwidth]{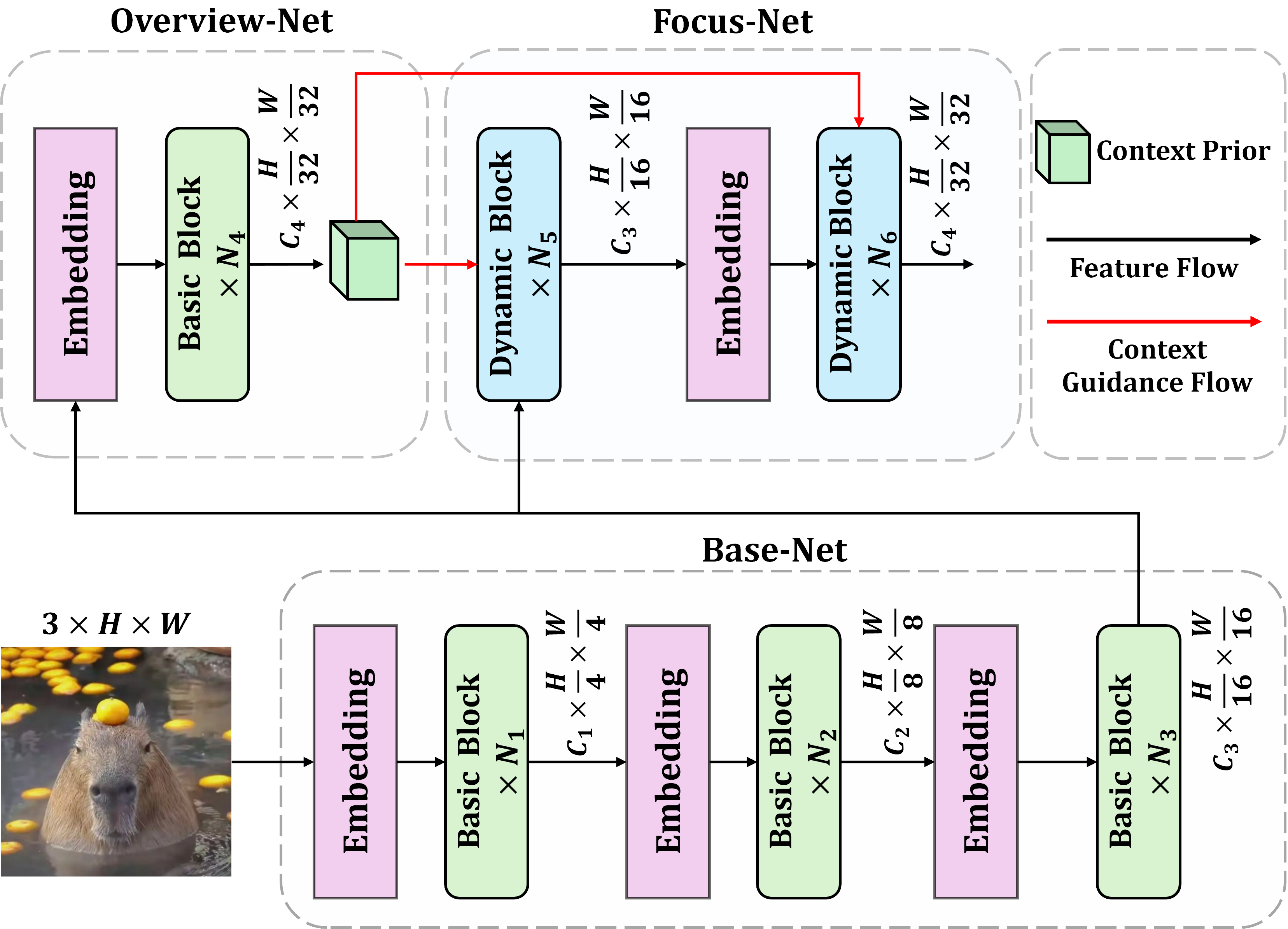}
\vspace{-6mm}
\caption{
The architecture of our OverLoCK network.\vspace{-4mm}
}
\label{fig:net}
\vspace{-4mm}
\end{figure}

\textbf{Overview}. Driven by the ``Overview-first-Look-Closely-next" mechanism of the human vision system~\cite{gilbert2007brain,li2014understanding}, we propose a deep-stage decomposition strategy (DDS) which, in contrast to classic hierarchical architectures, decomposes the network into three distinct sub-networks: Base-Net, Overview-Net, and Focus-Net. As shown in Figure~\ref{fig:net}, 
the Base-Net produces a mid-level feature map by progressively downsampling an input image to $\frac{H}{16} \times \frac{W}{16}$ via three embedding layers.
%implemented as strided 3$\times$3 convolutions. 
This mid-level feature map is fed into both the lightweight Overview-Net as well as the deeper and more powerful Focus-Net. The Overview-Net quickly produces a semantically meaningful but low-quality overview feature map, that serves as an overall understanding of the input image, by immediately downsampling the mid-level feature map to $\frac{H}{32} \times \frac{W}{32}$. This overview feature map is in fact used as a feedback signal fused into all building blocks of the Focus-Net to provide overall contextual information. Thus it is called the \textit{context prior}. Finally, guided by the \textit{context prior}, the Focus-Net goes back to progressively refine the mid-level feature map while enlarging the receptive field to obtain more accurate and informative high-level representations. Note that two backbone networks actually ``live" in the above design, one created by cascading Base-Net and Overview-Net and the other by cascading Base-Net and Focus-Net. Each backbone consists of four stages defined by four embedding layers and their following network building blocks. 
%Stage 3 of the Base-Net+Focus-Net backbone includes both basic and dynamic blocks while Stage 4 of the same backbone only has dynamic blocks. 
Our DDS design minimizes the overhead by having one Base-Net ``serving" mid-level features for both backbones.
\par
During pre-training on ImageNet-1K, to achieve representation learning in both Focus-Net and Overview-Net, each of them is connected to its own classifier head and the same classification loss is imposed on both classifiers. When the pre-trained network is transferred to downstream tasks, we no longer apply auxiliary supervision signals to Overview-Net as it has already learned high-level representations during the pre-training stage. Besides, applying auxiliary supervision in dense prediction tasks makes the training process time-consuming. Focus-Net is always used to make predictions in classification tasks. In dense prediction tasks, we use features from Base-Net at $\frac{H}{4} \times \frac{W}{4}$ and $\frac{H}{8} \times \frac{W}{8}$ resolutions as well as features from Focus-Net at $\frac{H}{16} \times \frac{W}{16}$ and $\frac{H}{32} \times \frac{W}{32}$ resolutions to construct a feature pyramid. These four groups of features also correspond to Stages 1 to 4 of our proposed ConvNet backbone network.

\begin{figure}[!t]
\centering
\includegraphics[width=0.465\textwidth]{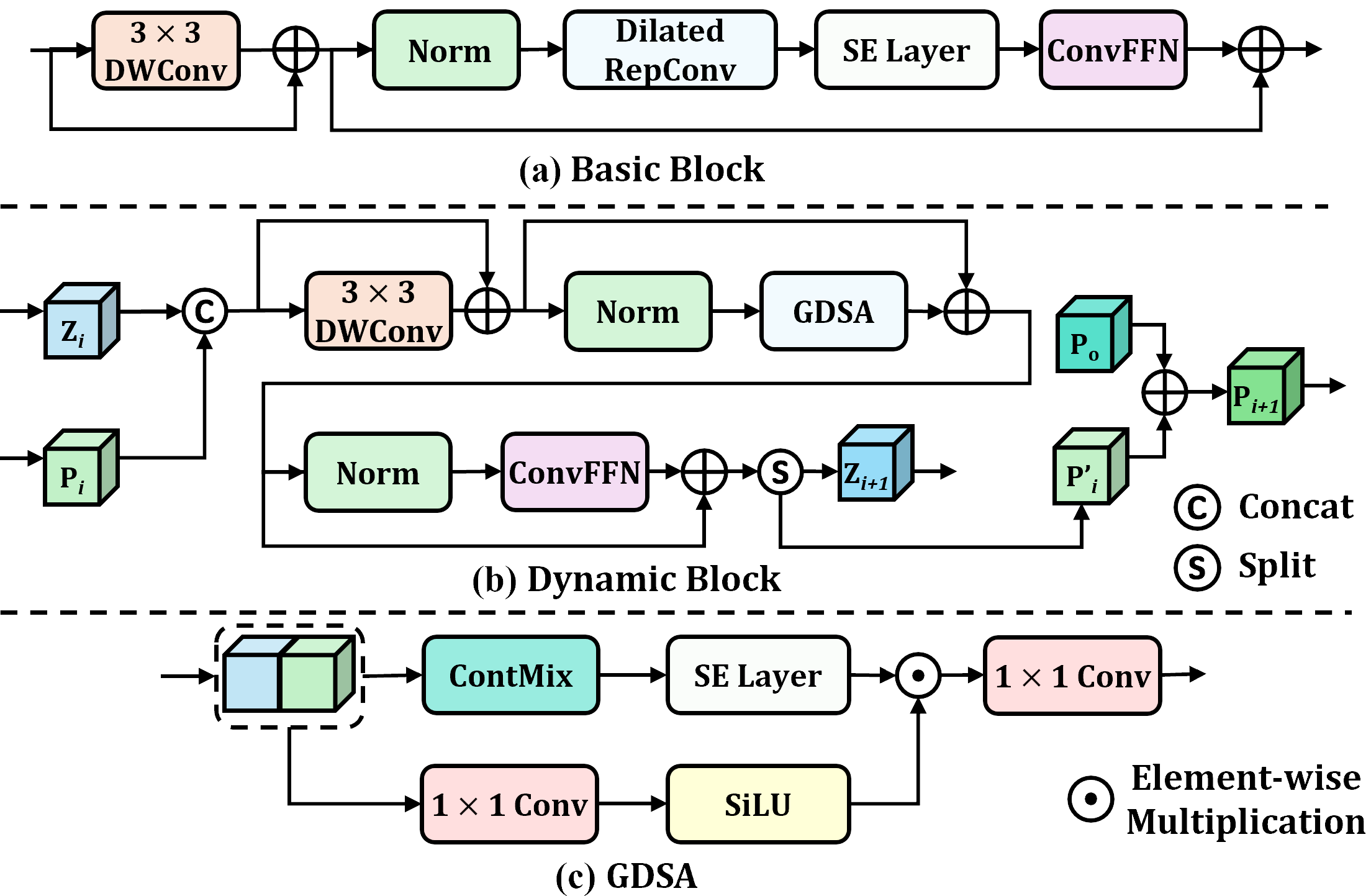}
\vspace{-3mm}
\caption{
Structures of network building blocks.
}
\label{fig:block}
\vspace{-6mm}
\end{figure}

\textbf{Base-Net and Overview-Net}. As shown in Figure \ref{fig:block} (a), we adopt the Basic Block as the building blocks of Base-Net and Overview-Net. The input feature is first fed into a residual 3$\times$3 DWConv to perform local perception. The output is then forwarded to a block consisting of a Layer Normalization \cite{lei2016layer} layer, a Dilated RepConv layer \cite{ding2023unireplknet}, a SE Layer \cite{jie2018squeeze}, and a ConvFFN \cite{wang2022pvt}.
\par
\textbf{Focus-Net}. As illustrated in Figure \ref{fig:block} (b), Focus-Net employs a more complex building block termed Dynamic Block mainly consisting of a residual 3$\times$3 DWConv, a Gated Dynamic Spatial Aggregator (GDSA), and a ConvFFN. The pipeline of GDSA is given in Figure \ref{fig:block} (c), where it uses the proposed ContMix (Section \ref{sec:dyconv}) as the core token mixer and additionally introduces a gated mechanism to eliminate contextual noise~\cite{gu2023mamba,li2023moganet,ma2024starnet}. Note that the Dynamic Blocks before the embedding layer in Focus-Net belong to Stage 3 of the Base-Net+Focus-Net backbone.

\par
\textbf{Context Flow}. There is a dynamic context flow within Focus-Net. The {\emph{context prior}} from Overview-Net not only provides guidance at both feature and kernel weight levels within Focus-Net, but also is updated within every block along the forward pass. 
% This process is elaborated as follows. 
%includes three stages: Context Input, Context Guidance, and Context Update.
% \par
Let us denote the {\emph{context prior}} and feature map at the entrance of the $i$-{th} block as $\mathbf{P}_i \in \mathbb{R}^{C_p \times H \times W}$ and $\mathbf{Z}_i \in \mathbb{R}^{C_z \times H \times W}$, respectively. $\mathbf{P}_i$ and $\mathbf{Z}_i$ are fused together through concatenation before being fed into the block (Figure~\ref{fig:block} (b)). Inside the block, feature-level guidance is achieved within GDSA by computing a dynamic gate to modulate the feature map using GDSA's input feature (Figure~\ref{fig:block} (c)), which is the result of applying a 1$\times$1 convolution followed by SiLU activation~\cite{elfwing2018sigmoid} to the aforementioned concatenated feature map. Subsequently, the dynamic gate is element-wise multiplied with the output of its parallel branch. On the other hand, to achieve weight-level guidance, the {\emph{context prior}} is injected into dynamic convolutions by utilizing $\mathbf{P}_i$ to compute dynamic kernel weights in ContMix, which will be elaborated in the next subsection. Before exiting the block, the fused feature map is split into $\mathbf{P'}_i \in \mathbb{R}^{C_p \times H \times W}$ and $\mathbf{Z}_{i+1} \in \mathbb{R}^{C_z \times H \times W}$, which can be regarded as the disentangled and updated {\emph{context prior}} and feature map. To prevent the {\emph{context prior}} from dilution, we add the initial {\emph{context prior}} $\mathbf{P}_o$ to $\mathbf{P'}_i$, i.e., $\mathbf{P}_{i+1} = \alpha \mathbf{P'}_i + \beta \mathbf{P}_o$, where $\alpha$ and $\beta$ are learnable scalars, both initialized to 1 before training.

%To save computation, at the beginning of the context flow, we reduce the number of channels in the original {\emph{context prior}} using a 1$\times$1 convolution layer, resulting in $\mathbf{P}_o \in \mathbb{R}^{C_p \times H \times W}$. Our experiments suggest that channel reduction by a factor of 4 yields the best performance. 
We perform channel reduction and spatial upsampling on the original {\emph{context prior}} to save computation and match the input resolution of Focus-Net, respectively. This results in the initial {\emph{context prior}} $\mathbf{P}_o$ of the context flow.

%Context Guidance includes feature-level guidance and weight-level guidance. As shown in Figure~\ref{fig:net}(c), we compute a dynamic gate using the concatenated feature map, aiming to recalibrate the feature representation. We also discuss different gate generation methods in experiments. On the other hand, we utilize the {\pcr{context prior}} to compute the dynamic weights in ContMix, injecting more accurate context perception into dynamic convolutions. Experiments suggest that this strategy yields improved performance.

\begin{figure*}[!t]
\centering
\includegraphics[width=0.85\textwidth]{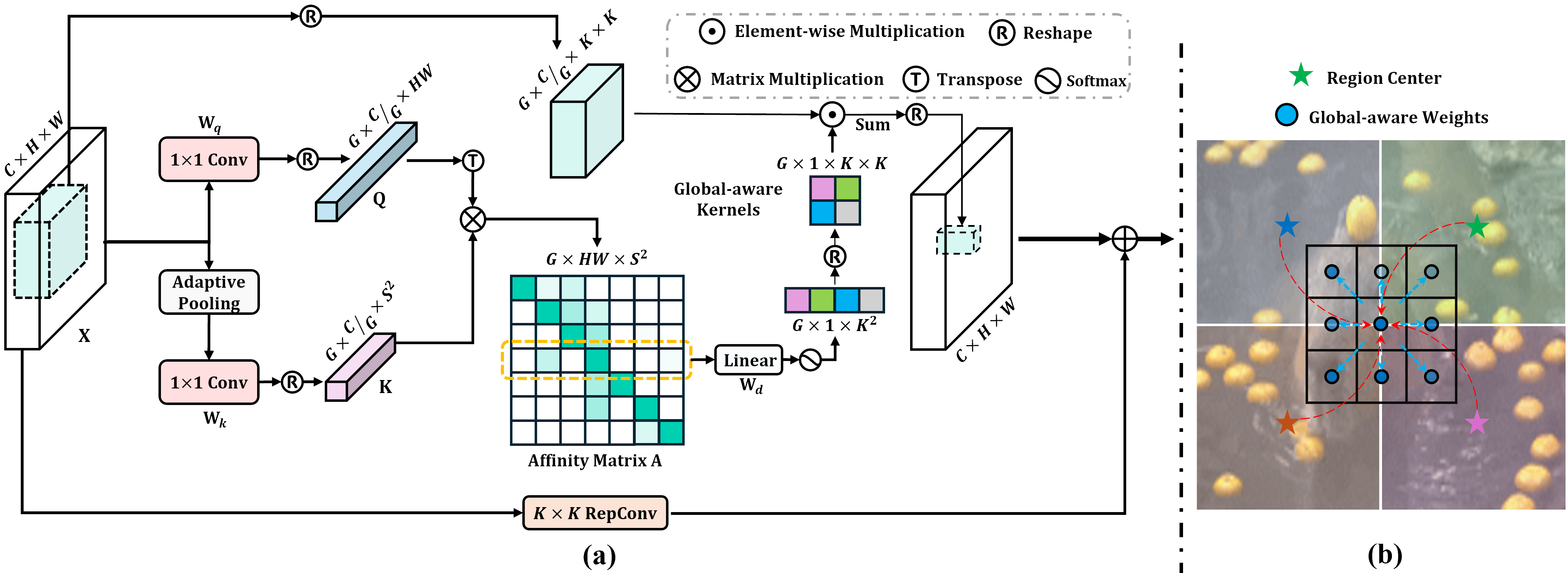}
% \vspace{-12pt}
\caption{
\textbf{(a)} A schematic diagram of our proposed dynamic convolution (ContMix). \textbf{(b)} An illustration of ContMix's ability in capturing long-range dependencies and preserving inductive biases.
% \vspace{-6mm}
}
\label{fig:dyconv}
\end{figure*}

\subsection{Dynamic Convolution with Context-Mixing}
\label{sec:dyconv}
In this section, we explore a solution that equips convolutions with the capability of long-range dependency modeling so that they can better handle varying input resolutions. Meanwhile, we still wish them to preserve strong inductive biases. To achieve these goals while fully leveraging the power of the context prior from Overview-Net, we propose a novel dynamic convolution that has a \textbf{Cont}ext-\textbf{Mix}ing ability, namely ContMix. Our key idea is to represent the relation between a token and its context using the set of affinity values between this individual token and all the tokens at a set of region centers in a feature map. These affinity values can then be aggregated to define token-wise dynamic convolution kernels in a learnable manner, thereby injecting contextual knowledge into every weight of the convolution kernels. Once such dynamic kernels have been applied to the feature map via sliding windows, every token in the feature map becomes modulated by the approximate global information gathered through the region centers. Thus long-range dependencies can be effectively modeled.
\par
\textbf{Token-wise Global Context Representation}. 
As shown in Figure \ref{fig:dyconv}, given an input feature map $\mathbf{X} \in \mathbb{R}^{C \times H \times W}$, we first transform it into two parts, namely, $\mathbf{Q} \in \mathbb{R}^{C \times HW} = \mathrm{Re}(\mathbf{W}_{q} \mathbf{X})$ and $\mathbf{K} \in \mathbb{R}^{C \times S^2} = \mathrm{Re}(\mathbf{W}_{k}\mathrm{Pool}(\mathbf{X}))$, where $\mathbf{W}_q$ and $\mathbf{W}_k$ denote 1$\times$1 convolutional layers, $\mathrm{Re}(\cdot)$ refers to the reshape operation, $\mathbf{K}$ represents the aggregation of $\mathbf{X}$ into $S \times S$ region centers via adaptive average pooling.
Next, we evenly divide the channels of $\mathbf{Q}$ and $\mathbf{K}$ into $G$ groups to obtain $\{\mathbf{Q^g}\}_{g=1}^G$ and $\{\mathbf{K^g}\}_{g=1}^G$ such that $\mathbf{Q^g} \in \mathbb{R}^{\frac{C}{G} \times HW}$ and $\mathbf{K^g} \in \mathbb{R}^{\frac{C}{G} \times S^2}$. Groups here are analogous to heads in multi-head attention~\cite{dosovitskiy2020image}. As every pair of $\mathbf{Q^g}$ and $\mathbf{K^g}$ have been flattened into 2D matrices, simple matrix multiplications between them computes $G$ affinity matrices $\{\mathbf{A^g}\}_{g=1}^G = \{\mathbf{Q^g}^{\mathrm{T}}\mathbf{K^g}\}_{g=1}^G$ where $\mathbf{A^g} \in \mathbb{R}^{HW \times S^{2}}$. The $i$-th row of affinity matrix $\mathbf{A^g}$, $\mathbf{A^g}_i$, holds the affinity values between the $i$-th token in $\mathbf{Q^g}$ and all tokens in $\mathbf{K^g}$.
\par
\begin{table}[b]
% \vspace{-20pt}
  \centering
  \caption{The configurations of OverLoCK variants.}
  \vspace{-10pt}
  \resizebox{0.49\textwidth}{!}{
    \begin{tabular}{ccccc}
    \toprule
    OverLoCK & \textit{Channels} & \textit{Blocks} & \textit{Kernel Sizes} & \textit{Groups} \\
    \midrule
    XT & $\left\{[56, 112, 256], [256], [256, 336]\right\}$ & $\left\{[2, 2, 3], [2], [6, 2]\right\}$ & $\left\{[17, 15, 13], [7], [13, 7]\right\}$ & $[4, 6]$ \\
    T & $\left\{[64, 128, 256], [512], [256, 512]\right\}$ & $\left\{[4, 4, 6], [2], [12, 2]\right\}$ & $\left\{[17, 15, 13], [7], [13, 7]\right\}$ & $[4, 8]$ \\
    S & $\left\{[64, 128, 320], [512], [320, 512]\right\}$ & $\left\{[6, 6, 8], [3], [16, 3]\right\}$ & $\left\{[17, 15, 13], [7], [13, 7]\right\}$ & $[5, 8]$ \\
    B & $\left\{[80, 160, 384], [576], [384, 576]\right\}$ & $\left\{[8, 8, 10], [4], [20, 4]\right\}$ & $\left\{[17, 15, 13], [7], [13, 7]\right\}$ & $[6, 9] $\\
    \bottomrule
    \end{tabular}%
    }
  \label{tab:model_variants}%
% \vspace{-12.5pt}
\end{table}%
\textbf{Token-wise Global Context Mixing}. 
To generate more robust feature representations, we define $G$ spatially varying $K \times K$ dynamic kernels. First, we use another learnable linear layer $\mathbf{W}_d \in \mathbb{R}^{S^{2} \times K^{2}}$ to aggregate the token-wise affinity values stored as matrix rows in every affinity matrix $\mathbf{A^g}$ by performing a matrix multiplication between $\mathbf{A^g}$ and $\mathbf{W}_d$. Note that all $G$ affinity matrices share the same $\mathbf{W}_d$ for saving computational efficiency. Then, a softmax function is employed to normalize the aggregated affinities. These two operations can be formulated as $\mathbf{D^g}=\mbox{softmax}(\mathbf{A^g}\mathbf{W}_d) \in \mathbb{R}^{HW \times K^{2}}$. Finally, every row of $\mathbf{D^g}$ can be reshaped into the target kernel shape to produce an input-dependent kernel at every token position. During the convolution operation, the channels of feature map $\mathbf{X}$ are also evenly divided into $G$ groups, and channels within the same group share the same dynamic kernel.
\par
\textbf{Implementation}. Our ContMix is a general plug-and-play module. In the Dynamic Block of our OverLoCK network, ContMix is customized as follows. The aforementioned $\mathbf{Q}$ and $\mathbf{K}$ matrices are computed using the channels of $\mathbf{X}$ corresponding to $\mathbf{Z}_i$ and $\mathbf{P}_i$ (the latest \textit{context prior}), respectively. This setting gives rise to better performance in comparison to computing both $\mathbf{Q}$ and $\mathbf{K}$ using the current fused feature $\mathbf{X}$. In addition, we empirically set $S$ to 7, ensuring that our ContMix enjoys linear-time complexity. Meanwhile, many previous works~\cite{ding2022scaling,liu2022more,ding2023unireplknet} suggest that combining large and small kernels can lead to better extraction of multi-scale features. Therefore, we allocate half of the groups in ContMix to large kernels and the remaining groups to small kernels, whose size is set to 5$\times$5 following previous works, enabling the modeling of long-range dependencies and local details using different kernels. We also employ a Dilated RepConv layer with $K\times K$ kernels to increase channel diversity.
%\vspace{-5pt}

\begin{table*}[t]
    \begin{minipage}[t]{0.64\linewidth}
      \centering
  \caption{A comparison of image classification performance on ImageNet-1K with 224$\times$224 resolution. \#F and \#P denote the FLOPs and number of Params of a model, respectively. \#T refers to model type, where ``C", ``T", ``M", and ``H" refer to ConvNet, Transformer, Mamba, and hybrid models, respectively.}
  \resizebox{0.99\textwidth}{!}{
    \begin{tabular}{lccccrlcccc}
\cmidrule{1-5}\cmidrule{7-11}    Method & \# T  & \# F (G) & \# P (M) & Acc. (\%) &       & Method & \# T  & \# F (G) & \# P (M) & Acc. (\%) \\
\cmidrule{1-5}\cmidrule{7-11}    PVTv2-B1\cite{wang2022pvt} & T     & 2.1   & 14    & 78.7  &       & Swin-S\cite{liu2021swin} & T     & 8.7   & 50    & 83.0  \\
    QuadTree-B-b1\cite{tang2022quadtree} & T     & 2.3   & 14    & 80.0  &       & PVTv2-B4\cite{wang2022pvt} & T     & 10.1  & 63    & 83.6  \\
    RegionViT-T\cite{chen2021regionvit} & T     & 2.4   & 14    & 80.4  &       & UniFormer-B\cite{li2022uniformer} & H     & 8.3   & 50    & 83.9  \\
    UniFormer-XS\cite{li2022uniformer} & H     & 2.0   & 17    & 82.0  &       & MaxViT-S\cite{tu2022maxvit} & T     & 11.7  & 69    & 84.5  \\
    CrossFormer-T\cite{wang2023crossformer++} & T     & 2.9   & 28    & 81.5  &       & NAT-S\cite{hassani2023neighborhood} & T     & 7.8   & 51    & 83.7  \\
    BiFormer-T\cite{zhu2023biformer} & T     & 2.2   & 13    & 81.4  &       & BiFormer-B\cite{zhu2023biformer} & T     & 9.8   & 57    & 84.3  \\
    NAT-M\cite{hassani2023neighborhood} & T     & 2.7   & 20    & 81.8  &       & VMamba-S\cite{liu2024vmamba} & M     & 8.7   & 50    & 83.6  \\
    GCViT-XT\cite{hatamizadeh2023global} & T     & 2.6   & 20    & 82.0  &       & ConvNeXt-S\cite{liu2022convnet} & C     & 8.7   & 50    & 83.1  \\
    ConvNeXt-N\cite{liu2022convnet} & C     & 2.7   & 16    & 80.9  &       & FocalNet-S\cite{yang2022focalnet} & C     & 8.7   & 50    & 83.5  \\
    VAN-B1\cite{guo2023visual} & C     & 2.5   & 14    & 81.1  &       & SLaK-S\cite{liu2022more} & C     & 9.8   & 55    & 83.8  \\
    Conv2Former-N\cite{HouConv2Former} & C     & 2.2   & 15    & 81.5  &       & RDNet-S\cite{kim2024densenets} & C     & 8.7   & 50    & 83.7  \\
    UniRepLKNet-N\cite{ding2023unireplknet}  & C     & 2.8   & 18    & 81.6  &       & InternImage-S\cite{wang2022internimage} & C     & 8.0   & 50    & 84.2  \\
    \rowcolor[rgb]{ .741,  .843,  .933} \textbf{OverLoCK-XT} & C     & 2.6   & 16    & $\mathbf{82.7 }$ & \cellcolor[rgb]{ 1,  1,  1} & \cellcolor[rgb]{ 1,  1,  1}InceptionNeXt-S\cite{yu2023inceptionnext} & \cellcolor[rgb]{ 1,  1,  1}C & \cellcolor[rgb]{ 1,  1,  1}8.4  & \cellcolor[rgb]{ 1,  1,  1}49  & \cellcolor[rgb]{ 1,  1,  1}83.5  \\
\cmidrule{1-5}          &       &       &       &       &       & PeLK-S\cite{chen2024pelk} & C     & 10.7  & 50    & 83.9  \\
    Swin-T\cite{liu2021swin} & T     & 4.5   & 28    & 81.3  &       & UniRepLKNet-S\cite{ding2023unireplknet} & C     & 9.1   & 56    & 83.9  \\
    PVTv2-B2\cite{wang2022pvt} & T     & 4.0   & 25    & 82.0  &       & MogaNet-B\cite{li2023moganet} & C     & 9.9   & 44    & 84.3  \\
    UniFormer-S\cite{li2022uniformer} & H     & 3.6   & 22    & 82.9  &       & \cellcolor[rgb]{ .741,  .843,  .933}\textbf{OverLoCK-S} & \cellcolor[rgb]{ .741,  .843,  .933}C & \cellcolor[rgb]{ .741,  .843,  .933}9.7  & \cellcolor[rgb]{ .741,  .843,  .933}56  & \cellcolor[rgb]{ .741,  .843,  .933}$\mathbf{84.8 }$ \\
\cmidrule{7-11}    MaxViT-T\cite{tu2022maxvit} & T     & 5.6   & 31    & 83.7  &       & Swin-B\cite{liu2021swin} & T     & 15.4  & 88    & 83.5  \\
    NAT-T\cite{hassani2023neighborhood} & T     & 4.3   & 28    & 83.2  &       & PVTv2-B5\cite{wang2022pvt} & T     & 11.8  & 82    & 83.8  \\
    BiFormer-S\cite{zhu2023biformer} & T     & 4.5   & 26    & 83.8  &       & NAT-B\cite{hassani2023neighborhood} & T     & 13.7  & 90    & 84.3  \\
    VMamba-T\cite{liu2024vmamba} & M     & 4.9   & 29    & 82.6  &       & MaxViT-B\cite{tu2022maxvit} & T     & 24.0  & 120   & 84.9  \\
    ConvNeXt-T\cite{liu2022convnet} & C     & 4.5   & 30    & 82.1  &       & VMamba-B\cite{liu2024vmamba} & M     & 15.4  & 89    & 83.9  \\
    FocalNet-T\cite{yang2022focalnet} & C     & 4.5   & 29    & 82.3  &       & ConvNeXt-B\cite{liu2022convnet} & C     & 15.4  & 89    & 83.8  \\
    SLaK-T\cite{liu2022more} & C     & 5.0   & 24    & 82.5  &       & FocalNet-B\cite{yang2022focalnet} & C     & 15.4  & 89    & 83.7  \\
    RDNet-T\cite{kim2024densenets} & C     & 5.0   & 30    & 82.8  &       & SLaK-B\cite{liu2022more} & C     & 17.1  & 95    & 84.0  \\
    InternImage-T\cite{wang2022internimage} & C     & 5.0   & 30    & 83.5  &       & RDNet-B\cite{kim2024densenets} & C     & 15.4  & 87    & 84.4  \\
    InceptionNeXt-T\cite{yu2023inceptionnext} & C     & 4.2   & 28    & 82.3  &       & InternImage-B\cite{wang2022internimage} & C     & 16.0  & 97    & 84.9  \\
    PeLK-T\cite{chen2024pelk} & C     & 5.6   & 29    & 82.6  &       & InceptionNeXt-B\cite{yu2023inceptionnext} & C     & 14.9  & 87    & 84.0  \\
    UniRepLKNet-T\cite{ding2023unireplknet} & C     & 4.9   & 31    & 83.2  &       & PeLK-B\cite{chen2024pelk} & C     & 18.3  & 89    & 84.2  \\
    MogaNet-S\cite{li2023moganet} & C     & 5.0   & 25    & 83.4  &       & MogaNet-L\cite{li2023moganet} & C     & 15.9  & 83    & 84.7  \\
    \rowcolor[rgb]{ .741,  .843,  .933} \textbf{OverLoCK-T} & C     & 5.5   & 33    & $\mathbf{84.2}$ & \cellcolor[rgb]{ 1,  1,  1} & \textbf{OverLoCK-B} & C     & 16.7  & 95    & $\mathbf{85.1 }$ \\
\cmidrule{1-5}\cmidrule{7-11}    
\end{tabular}%
}
\label{tab:cls}%
    \end{minipage}
    ~\begin{minipage}[t]{0.355\linewidth}
        \centering
        % \begin{table}
      \centering
      % \vspace{-7pt}
      \caption{A comparison of object detection and instance segmentation performance on the COCO dataset using Mask R-CNN. FLOPs are calculated for the 800$\times$1280 resolution.}
      % \vspace{-10pt}
          \resizebox{1\textwidth}{!}{
        \begin{tabular}{l|cc|cc|cc}
        \toprule
        \multirow{2}[4]{*}{Backbone} & \multirow{2}[4]{*}{\# F (G)} & \multirow{2}[4]{*}{\# P (M)} & \multicolumn{2}{c|}{1$\times$ Schedule} & \multicolumn{2}{c}{3$\times$ Schedule} \\
    \cmidrule{4-7}          &       &       & $AP^b$ & $AP^m$ & $AP^b$ & $AP^m$ \\
        \midrule
        Swin-T\cite{liu2021swin} & 267   & 48    & 42.7  & 39.3  & 46.0  & 41.6  \\
        PVTv2-B2\cite{wang2022pvt} & 309   & 45    & 45.3  & 41.2  & 47.8  & 43.1  \\
        UniFormer-S\cite{li2022uniformer} & 269   & 41    & 45.6  & 41.6  & 48.2  & 43.4  \\
        NAT-T\cite{hassani2023neighborhood} & 258   & 48    & -     & -     & 47.8  & 42.6  \\
        BiFormer-S\cite{zhu2023biformer} & 295   & 46    & 47.8  & 43.2  & -     & - \\
        VMamba-T\cite{liu2024vmamba} & 271   & 50    & 47.3  & 42.7  & 48.8  & 43.7  \\
        ConvNeXt-T\cite{liu2022convnet} & 262   & 48    & 44.2  & 40.1  & 46.2  & 41.7  \\
        FocalNet-T\cite{yang2022focalnet} & 268   & 49    & 46.1  & 41.5  & 48.0  & 42.9  \\
        InternImage-T\cite{wang2022internimage} & 270   & 49    & 47.2  & 42.5  & 49.1  & 43.7  \\
        RDNet-T\cite{kim2024densenets} & 278   & 43    & -     & -     & 47.3  & 42.2  \\
        MogaNet-S\cite{li2023moganet} & 272   & 45    & 46.7  & 42.2  & 48.5  & 43.1  \\
         \rowcolor[rgb]{ .741,  .843,  .933}\textbf{OverLoCK-T} & 281   & 52    & $\mathbf{48.3 }$ & $\mathbf{43.3 }$ & $\mathbf{49.6 }$ & $\mathbf{43.9 }$ \\
        \midrule
        Swin-S\cite{liu2021swin} & 354   & 69    & 44.8  & 40.9  & 48.2  & 43.2  \\
        PVTv2-B3\cite{wang2022pvt} & 397   & 65    & 47.0  & 42.5  & 48.4  & 43.2  \\
        UniFormer-B\cite{li2022uniformer} & 399   & 69    & 47.4  & 43.1  & 50.3  & 44.8  \\
        NAT-S\cite{hassani2023neighborhood} & 330   & 70    & -     & -     & 48.4  & 43.2  \\
        BiFormer-B\cite{zhu2023biformer} & 426   & 76    & 48.6  & 43.7  & -     & - \\
        VMamba-S\cite{liu2024vmamba} & 384   & 70    & 48.7  & 43.7  & 49.9  & 44.2  \\
        ConvNeXt-S\cite{liu2022convnet} & 348   & 70    & 45.4  & 41.8  & 47.9  & 42.9  \\
        FocalNet-S\cite{yang2022focalnet} & 365   & 72    & 48.3  & 43.1  & 49.3  & 43.8  \\
        InternImage-S\cite{wang2022internimage} & 340   & 69    & 47.8  & 43.3  & 49.7  & 44.5  \\
        MogaNet-B\cite{li2023moganet} & 373   & 63    & 47.9  & 43.2  & 50.3  & 44.4  \\
        \rowcolor[rgb]{ .741,  .843,  .933}\textbf{OverLoCK-S} & 366   & 75    & $\mathbf{49.4 }$ & $\mathbf{44.0 } $& $\mathbf{51.0 }$ & $\mathbf{45.0 }$ \\
        \midrule
        Swin-B\cite{liu2021swin} & 496   & 107   & 46.9  & 42.3  & 48.6  & 43.3  \\
        PVTv2-B5\cite{wang2022pvt} & 557   & 102   & 47.4  & 42.5  & 48.4  & 42.9  \\
        VMamba-B\cite{liu2024vmamba} & 485   & 108   & 49.2  & 43.9  & -     & - \\
        ConvNeXt-B\cite{liu2022convnet} & 486   & 108   & 47.0  & 42.7  & 48.5  & 43.5  \\
        FocalNet-B\cite{yang2022focalnet} & 507   & 111   & 49.0  & 43.5  & 49.8  & 44.1  \\
        InternImage-B\cite{wang2022internimage} & 501   & 115   & 48.8  & 44.0  & 50.3  & 44.8  \\
        MogaNet-L\cite{li2023moganet} & 495   & 102   & 49.4  & 44.1  & 50.5  & 44.5  \\
         \rowcolor[rgb]{ .741,  .843,  .933}\textbf{OverLoCK-B} & 511   & 114   & $\mathbf{49.9 }$ & $\mathbf{44.4 }$ & $\mathbf{51.4}$ & $\mathbf{45.3}$ \\
        \bottomrule
        \end{tabular}%
    }
      \label{tab:mask_det}%
      % \vspace{-4mm}
    % \end{table}%
    \end{minipage}
\vspace{-6mm}
\end{table*}

\subsection{Network Architecture}
\label{sec:net}
%\vspace{-5pt}
Our OverLoCK network has four architectural variants, including Extreme-Tiny (XT), Tiny (T), Small (S), and Base (B). As listed in Table~\ref{tab:model_variants}, we control the model size using four variables: \textit{Channels}, \textit{Blocks}, \textit{Kernel Sizes}, and \textit{Groups}. For instance, in OverLoCK-XT, \textit{Channels} = $\left\{ [56, 112, 256], [256], [256, 336] \right\}$, indicating that the channel counts in the three stages of Base-Net are [56, 112, 256], the channel count in Overview-Net is 256, and the channel counts in the two stages of Focus-Net are [256, 336]. \textit{Blocks} and \textit{Kernel Sizes} are similarly defined. Additionally, \textit{Groups}=[4, 6] indicates that the number of groups in the dynamic kernels of ContMix in the two stages of Focus-Net is 4 and 6, respectively.
%\vspace{-3mm}

\section{Experiments}
%\vspace{-4pt}
In this section, we present comprehensive experimental evaluations on various vision tasks, commencing with image classification. Then, we transfer the pre-trained models to downstream tasks, including object detection and semantic segmentation. Due to space constraints, we only report a subset of the results in this section, with additional experimental results provided in the \hyperref[sec:appendix_ab]{Appendix}.
%\vspace{-4pt}
\subsection{Image Classification}
\label{sec:cls}
\vspace{-4pt}
\textbf{Setup.} We conduct experiments on the ImageNet-1K dataset~\cite{deng2009imagenet} and adhere to the same experimental setting described in DeiT~\cite{touvron2021training} to ensure a fair comparison. Specifically, all models are trained for 300 epochs using the AdamW optimizer~\cite{loshchilov2017decoupled}. The stochastic depth rate~\cite{huang2016deep} is set to 0.1, 0.15, 0.4, and 0.5 for OverLoCK-XT, -T, -S, and -B models, respectively. All experiments are conducted on 8 NVIDIA H800 GPUs.
\par
\noindent\textbf{Results.} As shown in Table~\ref{tab:cls}, our pure ConvNet model achieves notable performance improvements over other competitors. For instance, OverLoCK-XT surpasses a strong Transformer-based model (BiFormer-T~\cite{zhu2023biformer}) and a recent large kernel ConvNet (UniRepLKNet-N \cite{ding2023unireplknet}) by significant 1.3\% and 1.1\% in Top-1 accuracy, respectively. For Tiny models, our OverLoCK-T also attains the best performance compared with other methods, achieving 84.2\% Top-1 accuracy, which improves upon MogaNet-S~\cite{li2023moganet} and PeLK-T~\cite{chen2024pelk} by 0.8\% and 1.6\% in Top-1 accuracy, respectively. When scaling up to larger models, our OverLoCK still maintains a significant advantage. Specifically, OverLoCK-S improves upon BiFormer-B and UniRepLKNet-S by notable 0.5\% and 0.9\% in Top-1 accuracy, respectively, with comparable computational complexity. Regarding the largest model, OverLoCK-B achieves an impressive 85.1\% Top-1 accuracy, outperforming MaxViT-B by 0.2\% in Top-1 accuracy with significantly lower computational complexity. Meanwhile, we evaluate the throughput of different models using a batch size of 128 on a single NVIDIA L40S GPU. Figure \ref{fig:acc_plot} demonstrates that our OverLoCK achieves an excellent tradeoff between speed and accuracy. For instance, OverLoCK-S surpasses MogaNet-B with over 100 imgs/s in throughput, while significantly increasing Top-1 accuracy from 84.3\% to 84.8\%. Similarly, OverLoCK-XT exceeds BiFormer-T by over 600 imgs/s in throughput, while remarkably improving Top-1 accuracy by 1.3\%. Overall, to the best of our knowledge, OverLoCK is the first pure ConvNet model to achieve such substantial performance gains over strong baselines on ImageNet-1K.
\begin{table*}[t]
    ~\begin{minipage}[t]{0.35\linewidth}
    % \begin{table}
      \centering
      \caption{A comparison of object detection and instance segmentation performance on the COCO dataset using Cascade Mask R-CNN. FLOPs are calculated for the 800$\times$1280 resolution.}
      % \vspace{-10pt}
        \resizebox{1\textwidth}{!}{
        \begin{tabular}{l|cc|cc}
        \toprule
        \multirow{2}[4]{*}{Backbone} & \multirow{2}[4]{*}{\# F (G)} & \multirow{2}[4]{*}{\# P (M)} & \multicolumn{2}{c}{3$\times$ Schedule} \\
    \cmidrule{4-5}          &       &       & $AP^b$ & $AP^m$ \\
        \midrule
        Swin-T\cite{liu2021swin} & 745   & 86    & 50.4  & 43.7  \\
        PVTv2-B2\cite{wang2022pvt} & 788   & 83    & 51.1  & - \\
        NAT-T\cite{hassani2023neighborhood} & 737   & 85    & 51.4  & 44.5  \\
        ConvNeXt-T\cite{liu2022convnet} & 741   & 86    & 50.4  & 43.7  \\
        HorNet-T\cite{rao2022hornet} & 730   & 80    & 51.7  & 44.8  \\
        RDNet-T\cite{kim2024densenets} & 757   & 81    & 51.6  & 44.6  \\
        PeLK-T\cite{chen2024pelk} & 770   & 86    & 51.4  & 44.6  \\
        UniRepLKNet-T\cite{ding2023unireplknet} & 749   & 89    & 51.8  & 44.9  \\
        MogaNet-S\cite{li2023moganet} & 750   & 78    & 51.6  & 45.1  \\
        \rowcolor[rgb]{ .741,  .843,  .933}\textbf{OverLoCK-T} & 760   & 90    & $\mathbf{52.4 }$ & $\mathbf{45.4 }$ \\
        \midrule
        Swin-S\cite{liu2021swin} & 838   & 107   & 51.8  & 44.7  \\
        NAT-S\cite{hassani2023neighborhood} & 809   & 108   & 52.0  & 44.9  \\
        ConvNeXt-S\cite{liu2022convnet} & 827   & 108   & 51.9  & 45.0  \\
        HorNet-S\cite{rao2022hornet} & 827   & 108   & 52.7  & 45.6  \\
        RDNet-S\cite{kim2024densenets} & 832   & 108   & 52.3  & 45.3  \\
        PeLK-S\cite{chen2024pelk} & 874   & 108   & 52.2  & 45.3  \\
        UniRepLKNet-S\cite{ding2023unireplknet} & 835   & 113   & 53.0  & 45.9  \\
        MogaNet-B\cite{li2023moganet} & 851   & 101   & 52.6  & 46.0  \\
        \rowcolor[rgb]{ .741,  .843,  .933}\textbf{OverLoCK-S} & 857   & 114   & $\mathbf{53.6 }$ & $\mathbf{46.4 }$ \\
        \midrule
        Swin-B\cite{liu2021swin} & 982   & 145   & 51.9  & 45.0  \\
        NAT-B\cite{hassani2023neighborhood} & 931   & 147   & 52.5  & 45.2  \\
        ConvNeXt-B\cite{liu2022convnet} & 964   & 146   & 52.7  & 45.6  \\
        HorNet-B\cite{rao2022hornet} & 969   & 144   & 53.3  & 46.1  \\
        RDNet-S\cite{kim2024densenets} & 971   & 144   & 52.3  & 45.3  \\
        PeLK-B\cite{chen2024pelk} & 1028  & 147   & 52.9  & 45.9  \\
        MogaNet-L\cite{li2023moganet} & 974   & 149   & 53.3  & 46.1  \\
         \rowcolor[rgb]{ .741,  .843,  .933}\textbf{OverLoCK-B} & 1008  & 154   & $\mathbf{53.9 }$ & $\mathbf{46.8 }$ \\
        \bottomrule
        \end{tabular}%
        }
      \label{tab:casc_det}%
    % \vspace{-4.5mm}
    % \end{table}%
    \end{minipage}
        % \vspace{-2.25em}
    ~\begin{minipage}[t]{0.2395\linewidth}
    % \begin{table}
      \centering
      \caption{Comparison of semantic segmentation performance on the ADE20K dataset. FLOPs are calculated for the 512$\times$2048 resolution.}
      \vspace{-10pt}
      \resizebox{1\textwidth}{!}{
        \begin{tabular}{l|ccc}
        \toprule
        \multirow{2}[3]{*}{Backbone} & \multicolumn{3}{c}{UperNet 160K} \\
    \cmidrule{2-4}          & F (G) & P (M) & mIoU \\
        \midrule
        Swin-T\cite{liu2021swin} & 945   & 60    & 44.5/45.8  \\
        UniFormer-S\cite{li2022uniformer} & 1008  & 52    & 47.6/48.5  \\
        NAT-T\cite{hassani2023neighborhood} & 934   & 58    & 47.1/48.4  \\
        BiFormer-S\cite{zhu2023biformer} & 1025  & 55    & 49.8/50.8  \\
        VMamba-T\cite{liu2024vmamba} & 949   & 62    & 48.0/48.8  \\
        ConvNeXt-T\cite{liu2022convnet} & 939   & 60    & 46.0/46.7  \\
        FocalNet-T\cite{yang2022focalnet} & 949   & 61    & 46.8/47.8  \\
        % HorNet-T & 926   & 52    & 48.1  & 48.9  \\
        SLaK-T\cite{liu2022more} & 936   & 65    & 47.6/----- \\
        RDNet-T\cite{kim2024densenets} & 961   & 58    & 47.6/48.6  \\
        InternImage-T\cite{wang2022internimage} & 944   & 59    & 47.9/48.1  \\
        InceptionNeXt-T\cite{yu2023inceptionnext} & 933   & 56    & 47.9/- \\
        PeLK-T\cite{chen2024pelk} & 970   & 62    & 48.1/----- \\
        UniRepLKNet-T\cite{ding2023unireplknet} & 946   & 61    & 48.6/49.1  \\
        MogaNet-S\cite{li2023moganet} & 946   & 55    & 49.2/-----  \\
        \rowcolor[rgb]{ .741,  .843,  .933}\textbf{OverLoCK-T} & 969   & 63    & $\mathbf{50.3}$/$\mathbf{50.8}$ \\
        \midrule
        Swin-S\cite{liu2021swin} & 1038  & 81    & 47.6/49.5  \\
        UniFormer-B\cite{li2022uniformer} & 1227  & 80    & 50.0/50.8  \\
        NAT-S\cite{hassani2023neighborhood} & 1071  & 82    & 48.0/49.5  \\
        BiFormer-B\cite{zhu2023biformer} & 1184  & 88    & 51.0/51.7  \\
        VMamba-S\cite{liu2024vmamba} & 1038  & 82    & 50.6/51.2  \\
        ConvNeXt-S\cite{liu2022convnet} & 1027  & 82    & 48.7/49.6  \\
        FocalNet-S\cite{yang2022focalnet} & 1044  & 84    & 49.1/50.1  \\
        % HorNet-S & 1030  & 81    & 49.2/49.8  \\
        SLaK-S\cite{liu2022more} & 1028  & 91    & 49.4/----- \\
        RDNet-S\cite{kim2024densenets} & 1040  & 86    & 48.7/49.8 \\
        InternImage-S\cite{wang2022internimage} & 1017  & 80    & 50.1/50.9  \\
        InceptionNeXt-S\cite{yu2023inceptionnext} & 1020  & 78    & 50.0/- \\
        PeLK-S\cite{chen2024pelk} & 1077  & 84    & 49.7/----- \\
        UniRepLKNet-S\cite{ding2023unireplknet} & 1036  & 86    & 50.5/51.0  \\
        MogaNet-B\cite{li2023moganet} & 1050  & 74    & 50.1/----- \\
        \rowcolor[rgb]{ .741,  .843,  .933}\textbf{OverLoCK-S} & 1051  & 85    & $\mathbf{51.3}$/$\mathbf{51.9}$ \\
        \midrule
        Swin-B\cite{liu2021swin} & 1188  & 121   & 48.1/49.7  \\
        NAT-B\cite{hassani2023neighborhood} & 1137  & 123   & 48.5/49.7  \\
        FocalNet-B\cite{yang2022focalnet} & 1192  & 126   & 50.5/51.4  \\
        % HorNet-B & 1174  & 121   & 50.0/50.5  \\
        SLaK-B\cite{liu2022more} & 1172  & 135   & 50.2/----- \\
        RDNet-B\cite{kim2024densenets} & 1187  & 127   & 49.6/50.5  \\
        InternImage-B\cite{yu2023inceptionnext} & 1185  & 128   & 50.8/51.3  \\
        InceptionNeXt-B\cite{yu2023inceptionnext} & 1159  & 115   & 50.6/----- \\
        PeLK-S\cite{chen2024pelk} & 1237  & 126   & 50.4/----- \\
        MogaNet-L\cite{li2023moganet} & 1176  & 113   & 50.9/----- \\
        \rowcolor[rgb]{ .741,  .843,  .933}\textbf{OverLoCK-B} & 1202  & 124   & $\mathbf{51.7}$/$\mathbf{52.3}$ \\
        \bottomrule
        \end{tabular}%
    }
      \label{tab:seg}%
    \end{minipage}
~\begin{minipage}[t]{0.4\linewidth}
    \centering
  \caption{A comprehensive roadmap that incrementally evolves a simple baseline to our OverLoCK-XT model. $\dagger$: The auxiliary loss for Overview-Net is only used in the classification task.}
    \resizebox{0.99\textwidth}{!}{
    \begin{tabular}{lcccc}
    \toprule
    Method & \# F (G) & \# P (M) & Top-1 (\%) & \multicolumn{1}{c}{mIoU (\%)} \\
    \midrule
    PlainNet & 2.5   & 15.7  & 76.3  &38.8  \\
    w/ Dilated RepConv & 2.5   & 15.7  & 76.6  &39.3  \\
    w/ SE & 2.5   & 16.0  & 77.1  &39.6  \\
    w/ Local Conv & 2.5   & 16.1  & 78.0  &40.2  \\
    FFN$\rightarrow$ConvFFN & 2.6   & 16.3  & 78.5  &41.1  \\
    \midrule
    Recurrent Model & 4.4   & 16.4  & 76.8  &39.5  \\
    \midrule
    \rowcolor{gray!10}DDS Model  & 2.6   & 15.6  & 79.0  &41.6  \\
    \rowcolor{gray!10}w/o feature feed & 2.7   & 16.3  & 78.2  &40.0  \\
    \rowcolor{gray!20}Static$\rightarrow$Dynamic & 2.6   & 15.8  & 80.0  &42.9  \\
    % \midrule
    % \rowcolor{gray!20}w/o DDS  & 2.6   & 16.4  & 79.4  &42.1 \\
    % \midrule
    \rowcolor{gray!30}w/ Aux Loss$^{\dagger}$  & 2.6   & 15.8  & 80.2  &43.1 \\
    \rowcolor{gray!40}w/ Initial Prior  & 2.6   & 15.8  & 80.4  &43.4 \\
    \rowcolor{gray!50}w/ Gate (OverLoCK-XT) & 2.6   & 16.4  & \textbf{80.8}  &\textbf{43.8}  \\
    \bottomrule
    \end{tabular}%
  \label{tab:roadmap}%
  }
  \vspace{10pt}
  \centering
  \caption{A comparison of dynamic token mixers}
  \vspace{-7.5pt}
    \resizebox{0.99\textwidth}{!}{
    \begin{tabular}{lcccc}
    \toprule
    \multicolumn{1}{l}{Method} & \# F (G) & \# P (M) & Top-1 (\%) & mIoU (\%) \\
    \midrule
    DyConv\cite{chen2020dynamic} & 1.7   & 12.5  & 77.5  & 36.3  \\
    ODConv\cite{li2022omni} & 1.7   & 13.8  & 77.9  & 36.7  \\
    Involution\cite{li2021involution} & 1.9   & 12.2  & 78.3  & 37.5  \\
    VOLO\cite{yuan2022volo}  & 2.3   & 14.1  & 75.5  & 35.6  \\
    DCNv3\cite{wang2022internimage} & 2.6   & 16.3  & 78.7  & 37.5  \\
    Shifted Window\cite{liu2021swin} & 2.2   & 13.6  & 78.4  & 37.4  \\
    Natten\cite{hassani2023neighborhood} & 2.2   & 13.6  & 78.6  & 38.1  \\
    \rowcolor[rgb]{ .741,  .843,  .933}ContMix (Ours) & 2.2   & 13.1  & \textbf{78.8} & \textbf{39.0} \\
    \bottomrule
    \end{tabular}%
    }
  \label{tab:mixers}%
    \end{minipage}
      \vspace{-15pt}
\end{table*}

\subsection{Object Detection and Instance Segmentation}
\vspace{-5pt}
\textbf{Setup.} We evaluate our network architecture on object detection and instance segmentation tasks using the COCO 2017 dataset~\cite{lin2014microsoft}. We employ both Mask R-CNN~\cite{he2017mask} and Cascade Mask R-CNN~\cite{cai2019cascade} frameworks, adopting the same experimental setting in Swin~\cite{liu2021swin}. The backbone networks are initially pre-trained on ImageNet-1K and subsequently fine-tuned for 12 epochs (1$\times$ schedule) and 36 epochs (3$\times$ schedule with multi-scale training).
\par
\noindent\textbf{Results.} As shown in Tables~\ref{tab:mask_det} and \ref{tab:casc_det}, OverLoCK demonstrates notable superiority over other methods. For instance, using the Mask R-CNN 1$\times$ schedule, OverLoCK-S surpasses BiFormer-B and MogaNet-B by 0.8\% and 1.5\% in AP$^b$, respectively. When using Cascade Mask R-CNN, OverLoCK-S improves upon PeLK-S and UniRepLKNet-S by 1.4\% and 0.6\% in AP$^b$, respectively. Notably, we observe an interesting phenomenon: \textit{\textbf{although ConvNet-based methods achieve comparable performance with Transformer-based methods on image classification tasks, there is a significant performance gap on detection tasks.}} For example, both MogaNet-B and BiFormer-B achieve Top-1 accuracy of 84.3\% on ImageNet-1K, but the former lags behind the latter on detection tasks. This validates our previous argument that ConvNet's fixed kernel size leads to limited receptive fields, resulting in performance degradation when using large input resolutions. Conversely, our OverLoCK effectively captures long-range dependencies even at large resolutions, resulting in excellent performance.
\subsection{Semantic Segmentation}
\label{sec:seg}
\vspace{-5pt}
\textbf{Setup.} We conduct experiments on semantic segmentation using the ADE20K dataset~\cite{zhou2017scene} with the UperNet framework~\cite{xiao2018unified}. For a fair comparison, we initialize all backbone networks with weights pre-trained on ImageNet-1K, following the same training settings as outlined in Swin~\cite{liu2021swin}.
\par
\noindent\textbf{Results.} Table~\ref{tab:seg} demonstrates that our OverLoCK achieves leading performance on semantic segmentation. For instance, OverLoCK-T outperforms MogaNet-S and UniRepLKNet-T by 1.1\% and 1.7\% in terms of mIoU, respectively, and surpasses VMamba-T, which emphasizes global modeling, by 2.3\% in mIoU. This advantage is consistently observed in both Small and Base models. Moreover, \textit{\textbf{we find that the issue of limited receptive fields also negatively impacts the performance of ConvNets on segmentation tasks}}, e.g., MogaNet-B lags behind BiFormer-B by 0.9\% despite having the same accuracy on classification. In contrast, our OverLoCK effectively alleviates this issue.
\vspace{-1mm}
\subsection{Ablation Studies}
\label{sec:ab_study}
\vspace{-5pt}
\textbf{Setup.} We conduct comprehensive ablation studies on image classification and semantic segmentation tasks to evaluate the effectiveness of individual components in OverLoCK. Specifically, we train each model variant on the ImageNet-1K dataset for 120 epochs following \cite{liu2022more,chen2024pelk} while keeping the remaining training settings consistent with those described in Section~\ref{sec:cls}. Subsequently, we fine-tune the pre-trained models on the ADE20K dataset for 80K iteration steps, with a batch size of 32 for faster training, while keeping the remaining settings identical to those outlined in Section~\ref{sec:seg}. Due to the page limit, more ablation studies are presented in the \hyperref[sec:appendix_ab]{Appendix}.
\par
\textbf{A detailed roadmap to our OverLoCK model}. First, we aim to develop a powerful baseline model using static large kernel convolutions. To this end, we first evaluate the performance of different components in the Basic Block (Figure \ref{fig:block} (a)). Specifically, we construct a hierarchical model using a vanilla convolutional layer followed by a vanilla FFN~\cite{dosovitskiy2020image} as the building block. The model consists of four stages with the number of blocks in every stage set to $[2, 2, 9, 4]$ and the number of channels in every stage set to $[56, 112, 304, 400]$. The kernel sizes of the four stages are consistent with the XT model. This model is denoted as ``PlainNet" and achieves a Top-1/mIoU of 76.3\%/38.8\%, as listed in Table~\ref{tab:roadmap}. Then, we convert the vanilla convolutional layer to the Dilated RepConv layer \cite{ding2023unireplknet}, referring to ``w/ Dilated RepConv" (Top-1/mIoU: 76.6\%/39.3\%). Next, we incrementally add an SE Layer (Top-1/mIoU: 77.1\%/39.6\%), a 3$\times$3 DWConv (Top-1/mIoU: 78.0\%/40.2\%), and replace the vanilla FFN with ConvFFN (Top-1/mIoU: 78.5\%/41.1\%). The resulting network is termed ``Baseline".
\par
Subsequently, we explore three strategies to inject top-down attention into this Baseline network. (1) Inspired by AbsViT \cite{shi2023top}, we construct a recurrent model by upsampling the output of Stage 4 and concatenating it with the input of Stage 3, termed ``Recurrent Model". However, this model's performance drops to 76.8\%/39.5\% with higher complexity, indicating that recurrent designs are unsuitable for modern ConvNet-based backbones. (2) We employ our proposed DDS to decompose the Baseline network into three inter-connected sub-networks. The outputs from both Base-Net and Overview-Net are concatenated and fed into Focus-Net. To ensure similar computational costs with the Baseline model, the numbers of blocks and channels in the three sub-networks are set to $[56, 112, 304], [400], [304, 400]$ and $[2, 2, 3], [2], [6, 2]$, respectively. This model, denoted as ``DDS Model", achieves a Top-1 accuracy of 79.0\%/41.6\%. (3) In the ``DDS Model", we only feed the projected output of Overview-Net into Focus-Net, without concatenating it with the output of Base-Net, the resulting model is termed ``w/o feature feed". This model decreases the performance, demonstrating the importance of feeding the output from Base-Net into Focus-Net.
\par
Finally, we evaluate the impact of weight-level context guidance by replacing every existing block in the Focus-Net with the proposed Dynamic Block, excluding the gate module, and ensuring that the context prior update flow does not use the initial \textit{context prior}. This modification maintains the same numbers of blocks and channels as our XT model, ensuring comparable computational complexity for a fair comparison. This model, termed ``Static $\rightarrow$ Dynamic", notably improves the Top-1/mIoU to 80.0\%/42.9\%. Next, to make Overview-Net produce semantically meaningful context features,  we use an auxiliary classification loss to supervise its output. The resulting model is termed ``w/ Aux Loss", which further improves the Top-1/mIoU by 0.2\%/0.2\%. Subsequently, we incorporate the initial \textit{context prior} into each dynamic block, as described in Section \ref{fig:dyconv}, to prevent the dilution of meaningful information within the \textit{context prior} during the updating process. This variant, labeled as ``w/ Initial Prior" enhances the Top-1/mIoU by 0.2\%/0.3\%. Lastly, we evaluate the impact of context-guided feature modulation by adding the gate module. This results in our XT model, which further boosts the Top-1/mIoU to 80.8\%/43.8\%. Summarily, our proposed method plays a vital role in significant performance improvements.
\par
% \vspace{-5pt}
\textbf{A comparison of token mixers}. To conduct a fair comparison of dynamic token mixers, we construct a Swin-like architecture~\cite{liu2021swin} by setting the numbers of blocks and channels in the four stages to $[2, 2, 6, 2]$ and $[64, 128, 256, 512]$, respectively, and employing non-overlapping patch embedding and a standard feed-forward network (FFN). We implement DyConv and ODConv in a separable convolution style~\cite{chollet2017xception} to ensure comparable computational complexity with other methods. Additionally, we set the kernel/window size to 7$\times$7 for all methods except for VOLO, where larger kernels incur significantly more parameters. From Table \ref{tab:mixers}, our \textbf{Cont}ext-\textbf{Mix}ing Dynamic Kernel (ContMix) achieves the best result on both image classification and semantic segmentation tasks. Notably, although ContMix exhibits similar performance as Natten and DCNv3 on classification tasks with low-resolution inputs, it demonstrates a clear advantage on semantic segmentation tasks with higher-resolution inputs. This is because ContMix captures long-range dependencies while preserving local inductive biases.

\section{Conclusion}
% \vspace{-5pt}
This paper proposes a biomimetic Deep-stage Decomposition (DDS) mechanism that injects semantically meaningful contexts into the intermediate stages of the network and also presents a novel dynamic convolution with context-mixing capacity, dubbed ContMix, which captures long-range dependencies while preserving strong inductive biases. By integrating these components, we propose a powerful, pure ConvNet-based vision backbone network, termed OverLoCK, achieving clearly superior performance compared to strong baselines.
% \section*{Acknowledgement}
% This work was supported by Hong Kong Research Grants Council under the Collaborative Research Fund (Project Number: HKU C7004-22G). %%%% Main Paper

% %%%%% WARNING: do not forget to delete the supplementary pages from your submission
\appendix
\clearpage
\maketitlesupplementary
% \title{\textcolor{cyan}{OverLoCK}: An \textcolor{cyan}{Over}view-first-\textcolor{cyan}{Lo}ok-\textcolor{cyan}{C}losely-next ConvNet\\ with Context-Mixing Dynamic \textcolor{cyan}{K}ernels}

% %%%%%%%%% AUTHORS - PLEASE UPDATE
% \author{Meng Lou 
% \and
% Yizhou Yu 
% \and
% School of Computing and Data Science, The University of Hong Kong \\
% {\tt\small loumeng@connect.hku.hk}, {\tt\small yizhouy@acm.org}
% }

% \newpage
% \counterwithin{table}{section}
% \counterwithin{figure}{section}
% \renewcommand{\thetable}{\thesection.\alph{table}}
% \renewcommand{\thefigure}{\thesection.\alph{figure}}
% \setcounter{page}{1}
\setcounter{table}{0}
\setcounter{figure}{0}
\renewcommand{\thetable}{\Alph{table}}
\renewcommand{\thefigure}{\Alph{figure}}

\section{More Ablation Studies}
\label{sec:appendix_ab}
On the basis of the training settings outlined in Section \textcolor{red}{4.4}, we additionally conduct a series of in-depth ablation experiments to meticulously examine the impact of every component in our proposed method.
% Section \ref{sec:ab_study} 4.4
\par
% Table \ref{tab:kernel_size}
\textbf{Impact of Kernel Sizes}. We compared the performance under various settings of kernel sizes, as outlined in Table \textcolor{red}{6} (the definition of the kernel size in our proposed method is given in Section \textcolor{red}{3.3}). The results indicate that the configuration $\left\{[17, 15, 13], [7], [13, 7]\right\}$ yields the optimal performance on both image classification and semantic segmentation tasks. Further enlarging the kernels does not lead to additional improvements.
% Section \ref{sec:net} 3.3
\begin{table}[htbp]
  \centering
  \caption{Ablation study of the kernel size setting.}
    \resizebox{0.49\textwidth}{!}{
    \begin{tabular}{lcccc}
    \toprule
    \multicolumn{1}{l}{Kernel Sizes} & \# F (G) & \# P (M) & Top-1 (\%) & mIoU (\%) \\
    \midrule
    \multicolumn{1}{l}{{$\left\{[19, 17, 15], [7], [15, 7]\right\}$}} & 2.8   & 16.5  & 80.7  & 43.8  \\
    \rowcolor[rgb]{ .867,  .922,  .969}\multicolumn{1}{l}{{$\left\{[17, 15, 13], [7], [13, 7]\right\}$}} & 2.6   & 16.4  & \textbf{80.8} & \textbf{43.8} \\
    \multicolumn{1}{l}{{$\left\{[13, 11, 9], [7], [9, 7]\right\}$}} & 2.6   & 16.3  & 80.5  & 43.5  \\
    \multicolumn{1}{l}{{$\left\{[9, 9, 7], [7], [7, 7]\right\}$}} & 2.6   & 16.1  & 80.6  & 43.3  \\
    {$\left\{[7, 7, 7], [7], [7, 7]\right\}$} & 2.5   & 16.1  & 80.4  & 43.1  \\
    \bottomrule
    \end{tabular}%
    }
  \label{tab:kernel_size}%
\end{table}%

\textbf{Impact of Stage Ratio}. The \textit{Stage Ratio} means the ratio between the number of blocks in the last stage of Base-Net and the number of blocks in the first stage of Focus-Net. In the default setting of the OverLoCK model, the stage ratio is 1:2 with the intention of allocating more network blocks to Focus-Net for extracting robust contextual information. 
%To avoid a significant increase in parameters, the number of layers in the last stage of the Overview-Net is kept consistent with the number of layers in the last stage of the Focus-Net. 
In this section, we investigate the impact of \textit{Stage Ratio}. Apart from the default setting of 1:2, we further set \textit{Stage Ratio} to 1:1 and 1:3 while maintaining the total number of network blocks constant. The results presented in Table~\ref{tab:stage_ratio} demonstrate that a \textit{Stage Ratio} of 1:2 yields the best outcomes. We posit that this is because a too small \textit{Stage Ratio} results in insufficient number of blocks in Focus-Net, thereby hindering the extraction of discriminative deep features. Conversely, an excessively large \textit{Stage Ratio} leads to a shortage of blocks in Base-Net, 
%causing a lack of shallow local details and weak context prior encompassing meaningful semantic information, 
thereby providing insufficient contextual guidance.

\begin{table}[htbp]
  \centering
  \caption{Ablation study of different stage ratio settings.}
      \resizebox{0.45\textwidth}{!}{
    \begin{tabular}{ccccc}
    \toprule
    \multicolumn{1}{l}{Stage Ratio} & \# F (G) & \# P (M) & Top-1 (\%) & mIoU (\%) \\
    \midrule
    1:1   & 2.7   & 16.1  & 80.4  & 42.9  \\
    \rowcolor[rgb]{ .867,  .922,  .969}1:2   & 2.6   & 16.4  & \textbf{80.8} & \textbf{43.8} \\
    1:3   & 2.7   & 15.9  & 80.6  & 43.6  \\
    \bottomrule
    \end{tabular}%
    }
  \label{tab:stage_ratio}%
\end{table}%

\textbf{Impact of Channel Reduction Factor}. In the default configuration of the OverLoCK model, we employ a 1$\times$1 convolution to reduce the number of output channels of Overview-Net by a factor of 4 and concatenate this result with the output of Base-Net before forwarding it to Focus-Net. We term this reduction as the \textit{Channel Reduction Factor (CRF)}. Therefore, the value of \textit{CRF} determines the number of channels in the \textit{context prior}, thereby influencing the guidance capability. In this regard, we investigate the effects of different \textit{CRF} settings. It is important to note that during the adjustment of \textit{CRF}, we also modify the number of channels of Focus-Net to maintain similar complexities across different model variants. The results in Table~\ref{tab:crf} demonstrate that \textit{CRF}=4 yields the optimal performance.

\begin{table}[htbp]
  \centering
   \caption{Ablation study of channel reduction factor settings.}
    \resizebox{0.4\textwidth}{!}{
    \begin{tabular}{ccccc}
    \toprule
    CRF & \# F (G) & \# P (M) & Top-1 (\%) & mIoU (\%) \\
    \midrule
    2     & 2.6   & 16.1  & 80.5  & 42.9  \\
    \rowcolor[rgb]{ .867,  .922,  .969}4     & 2.6   & 16.4  & \textbf{80.8} & \textbf{43.8} \\
    6     & 2.7   & 16.6  & 80.7  & 43.4  \\
    8     & 2.7   & 16.7  & 80.6  & 43.0  \\
    \bottomrule
    \end{tabular}%
    }
  \label{tab:crf}%
\end{table}%

\textbf{Impact of Auxiliary Loss}. To explore the effects of applying the auxiliary loss to Overview-Net, we adjust the weight of the auxiliary loss, drawing inspiration from prior research~\cite{zhao2017pyramid}. Given that the architectures of the models in this comparison are consistent, we opt not conduct further experiments on segmentation tasks for the sake of simplicity. The results presented in Table~\ref{tab:auxloss} indicate that the utilization of an auxiliary loss improves accuracy, while varying the weight of the auxiliary loss does not lead to a notable impact on performance. This observation aligns with findings in previous study \cite{zhao2017pyramid}.

\begin{table}[htbp]
  \centering
  \caption{Ablation study of auxiliary loss.}
    \resizebox{0.425\textwidth}{!}{
    \begin{tabular}{cccccc}
    \toprule
    Aux Loss Ratio & 0     & 0.2   & \cellcolor[rgb]{ .867,  .922,  .969}0.4   & 0.8   & 1.0 \\
    \midrule
    Top-1 (\%) & 80.4  & 80.7  & \cellcolor[rgb]{ .867,  .922,  .969}\textbf{80.8} & 80.7  & 80.7 \\
    \bottomrule
    \end{tabular}%
  \label{tab:auxloss}%
}
\end{table}%
% ``Baseline" represents the ContMix utilized in our OverLoCK model.
% Table~\ref{tab:roadmap} 6
\textbf{Effectiveness of our DDS-based Top-down Network}. To evaluate the effectiveness of the proposed DDS, we reconstruct our OverLoCK-XT model as a standard hierarchical network. To be specific, we eliminate the top-down attention mechanism by removing the Overview-Net while keeping the same types of layers in the Base-Net and Focus-Net. To maintain comparable complexity with other models, the number of channels and layers in the four stages are set to [64, 112, 256, 360] and [2, 2, 9, 4], respectively. This model is denoted as the ``Hierarchical Model". Additionally, we compare it with the ``Baseline" model in Table \textcolor{red}{6} which is a fully static ConvNet. As shown in Table~\ref{tab:dds_compare}, the ``Hierarchical Model" results in a noticeable performance drop, demonstrating the effectiveness of our DDS-based top-down context guidance. However, when compared with the ``Baseline" model, it still exhibits significant advantages, clearly indicating the superiority of our proposed dynamic convolution module.

\begin{table}[htbp]
  \centering
  \caption{Effectiveness of the proposed DDS-based top-down network.}
        \resizebox{0.45\textwidth}{!}{
    \begin{tabular}{lcccc}
    \toprule
    \multicolumn{1}{l}{Method} & \# F (G) & \# P (M) & Top-1 (\%) & mIoU (\%) \\
    \midrule
    \multicolumn{1}{l}{Baseline Model} & 2.6   & 16.3  & 78.5  & 41.1 \\
    \multicolumn{1}{l}{Hierarchical Model} & 2.7   & 16.2  & 79.2  & 41.9  \\
    \rowcolor[rgb]{ .867,  .922,  .969}OverLoCK-XT & 2.6   & 16.4  & \textbf{80.7} & \textbf{43.8} \\
    \bottomrule
    \end{tabular}%
    }
  \label{tab:dds_compare}%
\end{table}%

\par
\textbf{Ablation Study of ContMix}. We conduct a comprehensive comparison of various components within our proposed ContMix framework, as presented in Section \textcolor{red}{3.2}. As listed in Table~\ref{tab:ab_contmix}, we initially compute $\mathbf{Q}$ and $\mathbf{K}$ using the fused feature map instead of utilizing the channels of $\mathbf{X}$ corresponding to $\mathbf{Z}_i$ and $\mathbf{P}_i$ (the latest \textit{context prior}). This model variant, referred to as ``Fusion Affinity", results in a marginal performance decline. Subsequently, we interchange the features used to generate the $\mathbf{Q}$ and $\mathbf{K}$ matrices. This model, denoted as ``Reverse QK", also exhibits a decrease in performance. Furthermore, we individually eliminate the Softmax function (referred to as ``w/o Softmax"), remove the RepConv (referred to as ``w/o RepConv"), and substitute small kernels with large kernels (referred to as ``w/o Small Kernel"). These alterations decrease performance on both classification and segmentation tasks.
% Section \ref{sec:dyconv} 3.2
\begin{table}[htbp]
  \centering
  \caption{Ablation study of ContMix.}
    \resizebox{0.45\textwidth}{!}{
    \begin{tabular}{ccccc}
    \toprule
    \multicolumn{1}{l}{Method} & \# F (G) & \# P (M) & Top-1 (\%) & mIoU (\%) \\
    \midrule
    \rowcolor[rgb]{ .867,  .922,  .969}\multicolumn{1}{l}{Baseline} & 2.6   & 16.4  & \textbf{80.8} & \textbf{43.8} \\
    \multicolumn{1}{l}{Fusion Affinity} & 2.7   & 16.6  & 80.7  & 43.5  \\
    \multicolumn{1}{l}{Reverse QK} & 2.7   & 16.4  & 80.6  & 42.9  \\
    \multicolumn{1}{l}{w/o Softmax} & 2.6   & 16.4  & 80.5  & 43.5  \\
    \multicolumn{1}{l}{w/o RepConv} & 2.5   & 16.1  & 80.6  & 43.4  \\
    \multicolumn{1}{l}{w/o Small Kernel} & 2.8   & 16.6  & 80.7  & 43.3  \\
    \bottomrule
    \end{tabular}%
    }
  \label{tab:ab_contmix}%
\end{table}%

\begin{table}[htbp]
  \centering
  \caption{A comparison of image classification with 384$\times$384 inputs.}
\resizebox{0.4\textwidth}{!}{
    \begin{tabular}{lcccc}
    \toprule
    Method & Type  & \# F (G) & \# P (M) & Acc. (\%) \\
    \midrule
    Swin-B & T     & 47.1  & 88    & 84.5  \\
    MaxViT-B & T     & 74.2  & 120   & 85.7  \\
    ConvNeXt-B & C     & 45.2  & 88    & 85.1  \\
    InceptionNeXt-B & C     & 43.6  & 87    & 85.2  \\
    RDNet-L & C     & 101.9  & 186   & 85.8  \\
    PeLK-B-101 & C     & 68.3  & 90    & 85.8  \\
    \rowcolor[rgb]{ .867,  .922,  .969}\textbf{OverLoCK-B} & C     & 50.4  & 95    & \textbf{86.2} \\
    \bottomrule
    \end{tabular}%
}
  \label{tab:in1k-384}%
\end{table}%

\begin{table}[ht]
  \centering
  \caption{Robustness comparisons of different models.}
      \resizebox{0.45\textwidth}{!}{
    \begin{tabular}{lccccccc}
    \toprule
    Models & \# F (G) & \# P (G) & 1K & V2    & A     & R     & Sketch \\
    \midrule
    Swin-T  & 4.5   & 28    & 81.3  & 69.7  & 21.1  & 41.5  & 29.3  \\
    VMamba-T & 4.9   & 29    & 82.6  & 72.0  & 27.0  & 45.4  & 32.9  \\
    ConvNeXt-T & 4.5   & 29    & 82.1  & 72.5  & 24.2  & 47.2  & 33.8  \\
    HorNet-T & 4.0   & 22    & 82.8  & 72.3  & 26.6  & 46.6  & 34.1  \\
    SLaK-T & 5.0   & 30    & 82.5  & 72.0  & 30.0  & 45.3  & 32.4  \\
    NAT-T  & 4.3   & 28    & 83.2  & 72.2  & 33.0  & 44.9  & 31.9  \\
    RDNet-T & 5.0   & 24    & 82.8  & 72.9  & 27.7  & 49.0  & 37.0  \\
    UniRepLKNet-T & 4.9   & 25    & 83.2  & 72.8  & 34.8  & 49.4  & 36.9  \\
    MogaNet-S & 5.0   & 33    & 83.4  & 72.6  & 33.4  & 49.7  & 37.8  \\
    \rowcolor[rgb]{ .867,  .922,  .969}\textbf{OverLoCK-T} & 5.5   & 33    & \textbf{84.2} & \textbf{74.0} & \textbf{39.4} & \textbf{53.3} & \textbf{40.6} \\
    \midrule
    Swin-S & 8.7   & 50    & 83.0  & 72.0  & 32.5  & 45.2  & 32.3  \\
    VMamba-S & 8.7   & 50    & 83.6  & 73.2  & 33.2  & 49.4  & 37.0  \\
    ConvNeXt-S  & 8.7   & 50    & 83.1  & 72.5  & 31.3  & 49.6  & 37.1  \\
    HorNet-S  & 8.8   & 50    & 84.0  & 73.6  & 36.2  & 49.7  & 36.9  \\
    SLaK-S & 9.8   & 55    & 83.8  & 73.6  & 39.3  & 50.9  & 37.5  \\
    NAT-S  & 7.8   & 51    & 83.7  & 73.2  & 37.4  & 47.3  & 34.3  \\
    RDNet-S & 8.7   & 50    & 83.7  & 73.8  & 33.5  & 52.8  & 39.8  \\
    UniRepLKNet-S & 9.1   & 56    & 83.9  & 73.7  & 38.3  & 50.6  & 36.9  \\
    MogaNet-B & 9.9   & 44    & 84.3  & 74.3  & 40.4  & 50.1  & 38.6  \\
    \rowcolor[rgb]{ .867,  .922,  .969}\textbf{OverLoCK-S} & 9.7   & 56    & \textbf{84.8} & \textbf{74.9} & \textbf{45.0} & \textbf{57.2} & \textbf{45.8} \\
    \midrule
    Swin-B & 15.4  & 88    & 83.5  & 72.4  & 35.4  & 46.5  & 32.7  \\
    VMamba-B & 15.4  & 89    & 83.9  & 73.5  & 37.2  & 49.5  & 38.5  \\
    ConvNeXt-B  & 15.4  & 89    & 83.8  & 73.7  & 36.7  & 51.2  & 38.2  \\
    HorNet-B  & 15.6  & 87    & 84.3  & 73.9  & 39.9  & 51.2  & 38.1  \\
    SLaK-B & 17.1  & 95    & 84.0  & 74.0  & 41.6  & 50.8  & 38.5  \\
    NAT-B  & 13.7  & 90    & 84.3  & 74.1  & 41.4  & 49.7  & 36.6  \\
    RDNet-B & 15.4  & 87    & 84.4  & 74.2  & 38.1  & 52.7  & 40.1  \\
    MogaNet-L & 15.9  & 83    & 84.7  & 74.0  & 41.0  & 52.2  & 39.0  \\
    \rowcolor[rgb]{ .867,  .922,  .969}\textbf{OverLoCK-B} & 16.7  & 95    & \textbf{85.1} & \textbf{75.4} & \textbf{47.7} & \textbf{58.5} & \textbf{46.0} \\
    \bottomrule
    \end{tabular}%
    }
  \label{tab:robust_compare}%
\end{table}%

\begin{table*}[th]
  \centering
  \caption{Speed comparison among various models. Throughput (Thr.) is tested on a single NVIDIA L40S GPU with a batch size of 128 and an image size of 3$\times$224$\times$224.}
  \resizebox{0.875\textwidth}{!}{
    \begin{tabular}{lccccrlcccc}
\cmidrule{1-5}\cmidrule{7-11}    Method & \# F (G) & \# P (M) & Thr. (imgs/s) & Acc. (\%) &       & Method & \# F (G) & \# P (M) & Thr. (imgs/s) & Acc. (\%) \\
\cmidrule{1-5}\cmidrule{7-11}    Swin-T & 4.5   & 28    & 1324  & 81.3  &       & FocalNet-T & 4.5   & 29    & 1251  & 82.3  \\
    Swin-S & 8.7   & 50    & 812   & 83.0  &       & FocalNet-S & 8.7   & 50    & 777   & 83.5  \\
    Swin-B & 15.4  & 88    & 544   & 83.5  &       & FocalNet-B & 15.4  & 89    & 481   & 83.7  \\
\cmidrule{1-5}\cmidrule{7-11}    MaxViT-T & 5.6   & 31    & 683   & 83.7  &       & SLaK-T & 5.0   & 30    & 1126  & 82.5  \\
    MaxViT-S & 11.7  & 69    & 439   & 84.5  &       & SLaK-S & 9.8   & 55    & 747   & 83.8  \\
    MaxViT-B & 24.0  & 120   & 241   & 84.9  &       & SLaK-B & 17.1  & 95    & 478   & 83.7  \\
\cmidrule{1-5}\cmidrule{7-11}    NAT-M & 2.7   & 20    & 1740  & 81.8  &       & InternImage-T & 5.0   & 30    & 1084  & 83.5  \\
    NAT-T & 4.3   & 28    & 1287  & 83.2  &       & InternImage-S & 8.0   & 50    & 740   & 84.2  \\
    NAT-S & 7.8   & 51    & 823   & 83.7  &       & InternImage-B & 16.0  & 97    & 481   & 84.9  \\
\cmidrule{7-11}    NAT-B & 13.7  & 90    & 574   & 84.3  &       & UniRepLKNet-N & 2.8   & 18    & 1792  & 81.6  \\
\cmidrule{1-5}    BiFormer-T & 2.2   & 13    & 1103  & 81.4  &       & UniRepLKNet-T & 4.9   & 31    & 1094  & 83.2  \\
    BiFormer-S & 4.5   & 26    & 527   & 83.8  &       & UniRepLKNet-S & 9.1   & 56    & 707   & 83.9  \\
\cmidrule{7-11}    BiFormer-B & 9.8   & 57    & 341   & 84.3  &       & MogaNet-S & 5.0   & 25    & 766   & 83.4  \\
\cmidrule{1-5}    VMamba-T & 4.9   & 29    & 1179  & 82.6  &       & MogaNet-B & 9.9   & 44    & 373   & 84.3  \\
    VMamba-S & 8.7   & 50    & 596   & 83.6  &       & MogaNet-L & 15.9  & 83    & 282   & 84.7  \\
\cmidrule{7-11}    VMamba-B & 15.4  & 89    & 439   & 83.9  &       & \cellcolor[rgb]{ .867,  .922,  .969}\textbf{OverLoCK-XT} & \cellcolor[rgb]{ .867,  .922,  .969}2.6  & \cellcolor[rgb]{ .867,  .922,  .969}16  & \cellcolor[rgb]{ .867,  .922,  .969}1672 & \cellcolor[rgb]{ .867,  .922,  .969}\textbf{82.7} \\
\cmidrule{1-5}    ConvNeXt-T & 4.5   & 29    & 1507  & 82.1  &       & \cellcolor[rgb]{ .867,  .922,  .969}\textbf{OverLoCK-T} & \cellcolor[rgb]{ .867,  .922,  .969}5.5  & \cellcolor[rgb]{ .867,  .922,  .969}33  & \cellcolor[rgb]{ .867,  .922,  .969}810 & \cellcolor[rgb]{ .867,  .922,  .969}\textbf{84.2} \\
    ConvNeXt-S & 8.7   & 50    & 926   & 83.1  &       & \cellcolor[rgb]{ .867,  .922,  .969}\textbf{OverLoCK-S} & \cellcolor[rgb]{ .867,  .922,  .969}9.7  & \cellcolor[rgb]{ .867,  .922,  .969}56  & \cellcolor[rgb]{ .867,  .922,  .969}480 & \cellcolor[rgb]{ .867,  .922,  .969}\textbf{84.8} \\
    ConvNeXt-B & 15.4  & 89    & 608   & 83.8  &       & \cellcolor[rgb]{ .867,  .922,  .969}\textbf{OverLoCK-B} & \cellcolor[rgb]{ .867,  .922,  .969}16.7  & \cellcolor[rgb]{ .867,  .922,  .969}95  & \cellcolor[rgb]{ .867,  .922,  .969}306 & \cellcolor[rgb]{ .867,  .922,  .969}\textbf{85.1} \\
\cmidrule{1-5}\cmidrule{7-11}    
\end{tabular}%
  }
  \label{tab:speed}%
\end{table*}%

\section{Additional Experiments on Image Classification}
\subsection{Large Resolution Evaluation}
Following previous works \cite{liu2022convnet,yu2023inceptionnext,kim2024densenets}, we further investigate the image classification performance on the ImageNet-1K dataset at a higher resolution (i.e., 384$\times$384). Specifically, we pre-train the base model on 224$\times$224 inputs and then fine-tune it on 384$\times$384 inputs for 30 epochs. As shown in Table \ref{tab:in1k-384}, our OverLock-B model achieves superior performance under high-resolution input conditions. Notably, OverLock-B surpasses MaxViT-B by 0.5\% in Top-1 accuracy while reducing the parameter count by over one-third. Compared to PeLK-B, a large kernel ConvNet, our method also demonstrates significant improvements. These results further validate the robustness of our proposed method in handling large-resolution inputs.

\subsection{Robustness Evaluation}
We further assess the robustness of our models using the ImageNet out-of-distribution (OOD) benchmarks, including ImageNet-V2~\cite{imagenetv2}, ImageNet-A~\cite{imagenet-A}, ImageNet-R~\cite{imagenet-R}, and ImageNet-Sketch~\cite{ImageNetSketch}. As shown in Table~\ref{tab:robust_compare}, our method demonstrates excellent robustness on different datasets, outperforming representative ConvNets, Vision Transformers, and Vision Mamba. Notably, although OverLoCK-B improves over MogaNet-L by 0.4\% in Top-1 accuracy on ImageNet-1K, it achieves significant gains on OOD datasets, with improvements of 1.4\% on ImageNet-V2, 6.7\% on ImageNet-A, 6.3\% on ImageNet-R, and 6.8\% on ImageNet-Sketch. These results showcase the strong robustness of our pure ConvNet.

\begin{figure}[thb]
    \centering
    \includegraphics[width=0.325\textwidth]{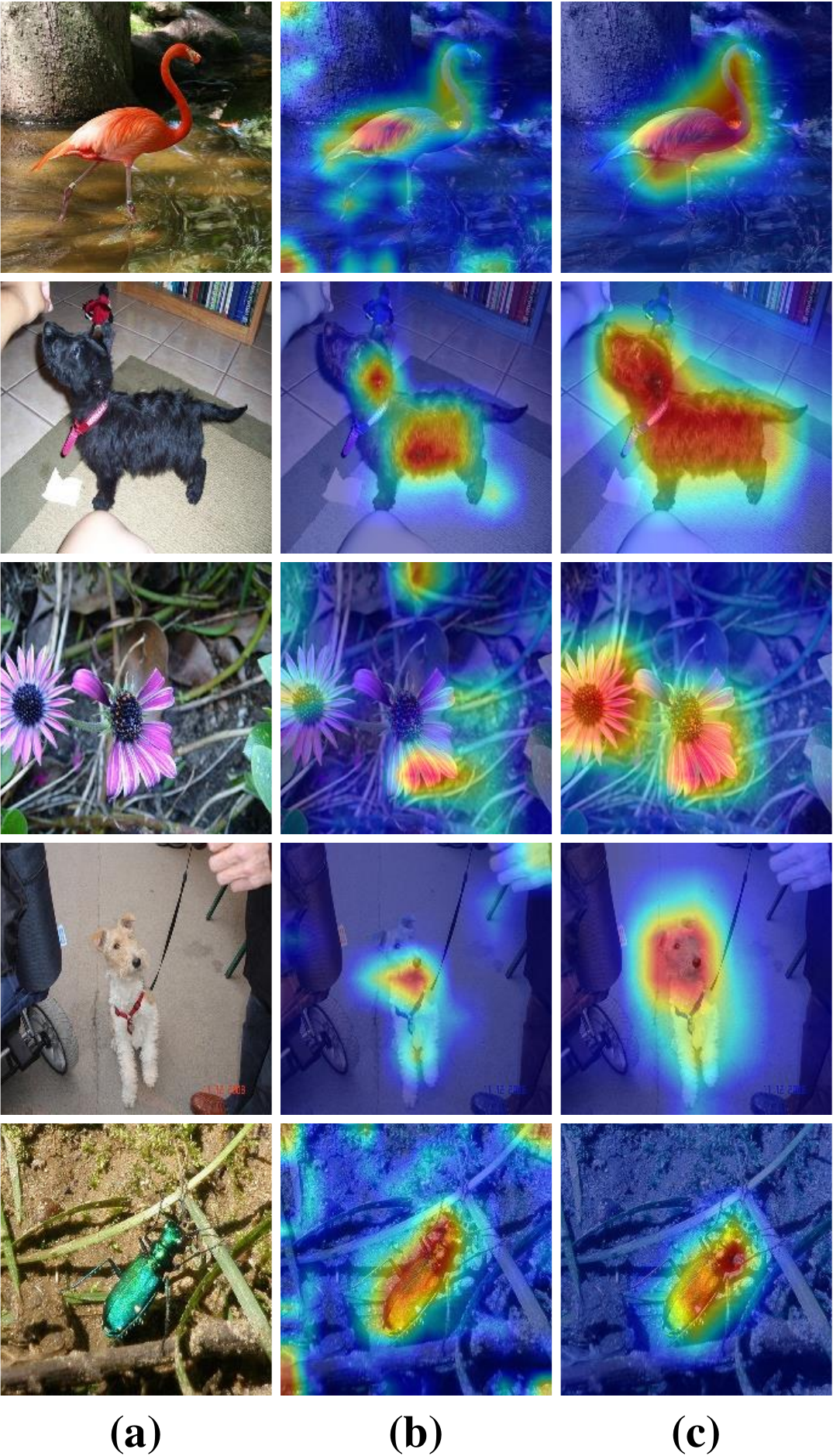}
    \caption{Class activation maps of the proposed OverLoCK network. (a), (b), and (c) show the input images, class activation maps of Overview-Net, and class activation maps of Focus-Net, respectively.}
    \label{fig:cam}
\end{figure}

\begin{figure}[th]
    \centering
    \includegraphics[width=0.47\textwidth]{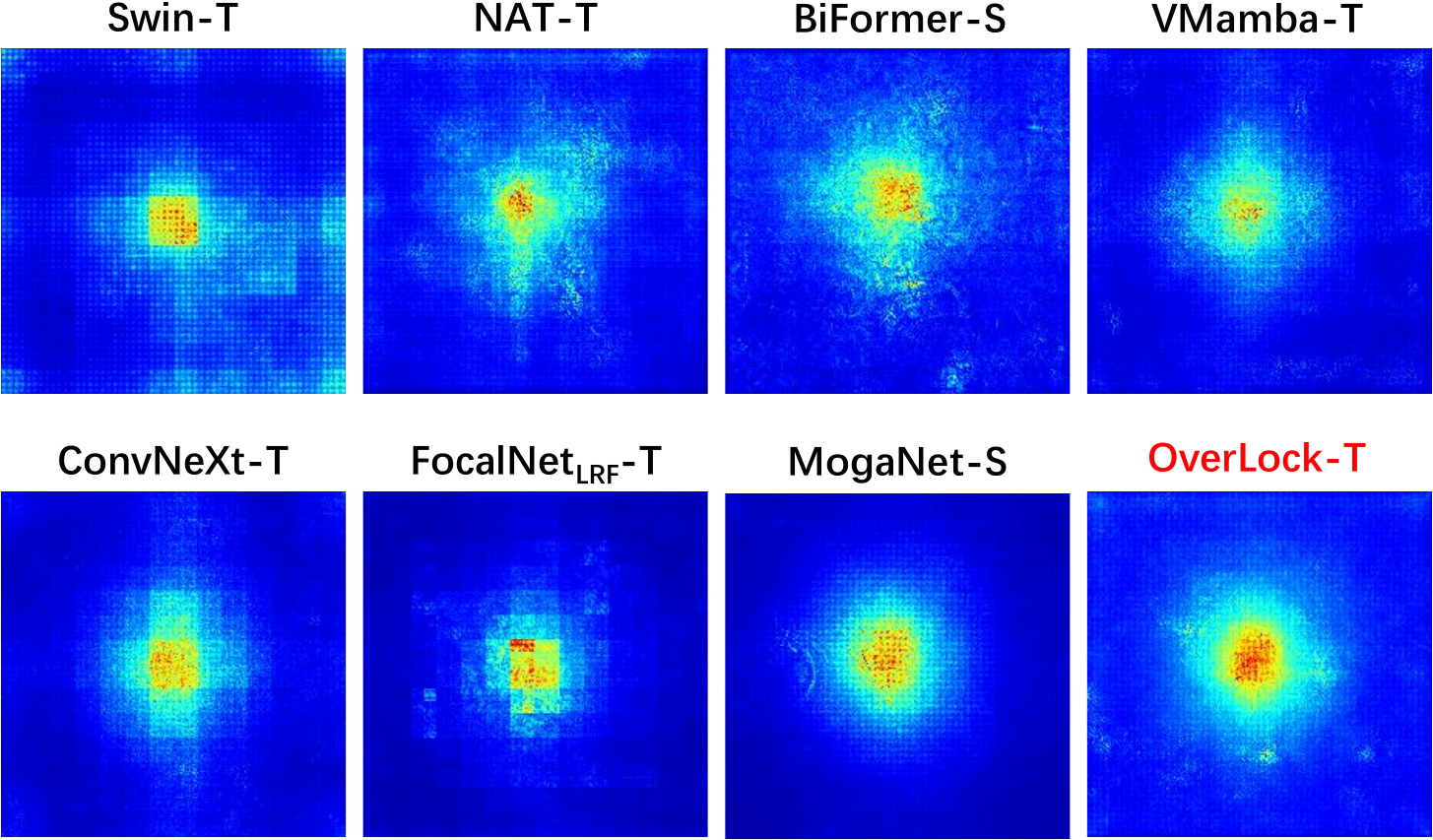}
    \caption{Comparison of ERF among various models.}
    \label{fig:erf}
\end{figure}

% \vspace{-8.75pt}
\section{Speed Analysis}
% \vspace{-7.5pt}
% Figure~\ref{fig:acc_plot} I
% Table~\ref{tab:speed} I
We provide a comparison of speed-accuracy trade-off in Figure \textcolor{red}{1}. More details are listed in Table \textcolor{red}{1}, where an OverLoCK variant often achieves faster speed and higher accuracy simultaneously than a larger variant of another network, demonstrating an excellent trade-off between speed and accuracy. For instance, OverLoCK-XT achieves 1672 imgs/s in throughput, improving upon Swin-T by over 300 imgs/s, while significantly enhancing Top-1 accuracy by 1.4\%. Also, OverLoCK-T achieves about 200 imgs/s improvement in throughput compared to ConvNeXt-B while achieving better performance at the cost of only around one-third of the FLOPS. When compared to more advanced models, OverLoCK still exhibits significant advantages. For example, OverLoCK-S surpasses MogaNet-B by over 100 imgs/s in throughput while increasing Top-1 accuracy from 84.3\% to 84.8\%. Likewise, OverLoCK-XT surpasses BiFormer-T by over 600 imgs/s in throughput while remarkably improving Top-1 accuracy by 1.3\%.

\section{Visualization Analysis}
\subsection{Effect of Context Guidance}
To visually understand the effect of context guidance, we separately visualize the class activation maps generated by Overview-Net and Focus-Net in OverLoCK-T using Grad-CAM~\cite{selvaraju2020grad} for the ImageNet-1K validation set. As shown in Figure~\ref{fig:cam}, Overview-Net first produces a coarse localization of an object, and when this signal is used as the top-down guidance for Focus-Net, the object's location and shape becomes more accurate.

\subsection{Effective Receptive Field Analysis}
To visually demonstrate the representation capacity of OverLoCK, we compare the Effective Receptive Field (ERF)~\cite{luo2016understanding} of our OverLoCK-T with that of other representative models with comparable complexity. The visualizations are generated using over 300 randomly sampled images with a resolution of 224$\times$224 from the ImageNet-1K validation set. As shown in Figure~\ref{fig:erf}, our model not only produces global responses but also exhibits significant local sensitivity, indicating that OverLoCK can effectively model both global and local contexts simultaneously.
% \vspace{-6.6pt}
% \section{Code}
% % \vspace{-6.6pt}
% We provide the pseudo code of ContMix in Algorithm~\ref{alg:dyconv}.

% \section{Rationale}
% \label{sec:rationale}
% % 
% Having the supplementary compiled together with the main paper means that:
% % 
% \begin{itemize}
% \item The supplementary can back-reference sections of the main paper, for example, we can refer to \cref{sec:intro};
% \item The main paper can forward reference sub-sections within the supplementary explicitly (e.g. referring to a particular experiment); 
% \item When submitted to arXiv, the supplementary will already included at the end of the paper.
% \end{itemize}
% % 
% To split the supplementary pages from the main paper, you can use \href{https://support.apple.com/en-ca/guide/preview/prvw11793/mac#:~:text=Delete%20a%20page%20from%20a,or%20choose%20Edit%20%3E%20Delete).}{Preview (on macOS)}, \href{https://www.adobe.com/acrobat/how-to/delete-pages-from-pdf.html#:~:text=Choose%20%E2%80%9CTools%E2%80%9D%20%3E%20%E2%80%9COrganize,or%20pages%20from%20the%20file.}{Adobe Acrobat} (on all OSs), as well as \href{https://superuser.com/questions/517986/is-it-possible-to-delete-some-pages-of-a-pdf-document}{command line tools}.

% \clearpage
{
\small
\bibliographystyle{ieeenat_fullname}
\bibliography{ref}
}

% %%%%%%%
% \appendix
% \input{sec/appendix}

% {
% \small
% \bibliographystyle{ieeenat_fullname}
% % \bibliographystyle{unsrt}
% \bibliography{ref}
% }
% %%%%%%%

\end{document}